\documentclass[10pt,journal,compsoc]{IEEEtran}

\ifCLASSOPTIONcompsoc
  \usepackage[nocompress]{cite}
\else
  \usepackage{cite}
\fi

\usepackage{amsmath}
\usepackage{amssymb}
\usepackage{float}
\usepackage{graphicx}
\usepackage{multirow}
\usepackage{longtable,booktabs}
\usepackage{xcolor,colortbl} % add color to table

\usepackage{algorithm}
\usepackage[noend]{algpseudocode}
\usepackage[pagebackref=true,breaklinks=true,letterpaper=true,colorlinks,bookmarks=false]{hyperref}
\usepackage{soul}
\usepackage{subfig}

\usepackage{cuted}
\usepackage{capt-of}

\DeclareMathOperator*{\argmin}{arg\,min}
\renewcommand{\vec}[1]{\boldsymbol{#1}}

\newcommand{\T}{\mathcal{T}}
\newcommand{\F}{\mathcal{F}}
\newcommand{\x}{\vec{x}}
\newcommand{\s}{\vec{s}}

\def\eg{\emph{e.g}.} \def\Eg{\emph{E.g}.}
\def\ie{\emph{i.e}.}

\def\etal{\emph{et al}.}

\makeatletter
\newcommand{\thickhline}{%
	\noalign {\ifnum 0=`}\fi \hrule height 1pt
	\futurelet \reserved@a \@xhline
}
\makeatother

\usepackage{scalerel}
\usepackage{tikz}
\usetikzlibrary{svg.path}

\definecolor{orcidlogocol}{HTML}{A6CE39}
\tikzset{
	orcidlogo/.pic={
		\fill[orcidlogocol] svg{M256,128c0,70.7-57.3,128-128,128C57.3,256,0,198.7,0,128C0,57.3,57.3,0,128,0C198.7,0,256,57.3,256,128z};
		\fill[white] svg{M86.3,186.2H70.9V79.1h15.4v48.4V186.2z}
		svg{M108.9,79.1h41.6c39.6,0,57,28.3,57,53.6c0,27.5-21.5,53.6-56.8,53.6h-41.8V79.1z M124.3,172.4h24.5c34.9,0,42.9-26.5,42.9-39.7c0-21.5-13.7-39.7-43.7-39.7h-23.7V172.4z}
		svg{M88.7,56.8c0,5.5-4.5,10.1-10.1,10.1c-5.6,0-10.1-4.6-10.1-10.1c0-5.6,4.5-10.1,10.1-10.1C84.2,46.7,88.7,51.3,88.7,56.8z};
	}
}

\newcommand\orcidicon[1]{\href{https://orcid.org/#1}{\mbox{\scalerel*{
				\begin{tikzpicture}[yscale=-1,transform shape]
				\pic{orcidlogo};
				\end{tikzpicture}
			}{|}}}}

\begin{document}

% title string
\def\titleStr{End-to-end Full Projector Compensation}

\title{\titleStr}

% authors
\author{Bingyao Huang\orcidicon{0000-0002-8647-5730},
        Tao Sun\orcidicon{0000-0002-6926-0543},
        and~Haibin Ling\orcidicon{0000-0003-4094-8413}
\IEEEcompsocitemizethanks{
  \IEEEcompsocthanksitem B. Huang, T. Sun and H. Ling are with Department of Computer Science, Stony Brook University, Stony Brook, NY 11794, USA. (E-mail: bihuang@cs.stonybrook.edu; tao@cs.stonybrook.edu; hling@cs.stonybrook.edu)\protect\\
%   \IEEEcompsocthanksitem T. Sun is with Department of Computer Science, Stony Brook University, Stony Brook, NY 11794, USA. (E-mail: tao@cs.stonybrook.edu)\protect\\
%   \IEEEcompsocthanksitem H. Ling is with Department of Computer Science, Stony Brook University, Stony Brook, NY 11794, USA. (E-mail: hling@cs.stonybrook.edu)\protect\\
%  \IEEEcompsocthanksitem  Preliminary versions of this work have appeared in CVPR 2019 \cite{huang2019compennet} and ICCV 2019 \cite{huang2019compennet++}.
}
\thanks{Copyright~\copyright~2021 IEEE.  Personal use of this material is permitted.  Permission from IEEE must be obtained for all other uses, in any current or future media, including reprinting/republishing this material for advertising or promotional purposes, creating new collective works, for resale or redistribution to servers or lists, or reuse of any copyrighted component of this work in other works.}}

% headers
\markboth{ACCEPTED BY IEEE TRANSACTIONS ON PATTERN ANALYSIS AND MACHINE INTELLIGENCE}%
{Huang \MakeLowercase{\emph{et al.}}: \titleStr}

\IEEEtitleabstractindextext{%
\begin{abstract}
  Full projector compensation aims to modify a projector input image to compensate for both geometric and photometric disturbance of the projection surface. Traditional methods usually solve the two parts separately and may suffer from suboptimal solutions.
  In this paper, we propose the first end-to-end differentiable solution, named CompenNeSt++, to solve the two problems jointly.
  First, we propose a novel geometric correction subnet, named WarpingNet, which is designed with a cascaded coarse-to-fine structure to learn the sampling grid directly from sampling images. Second, we propose a novel photometric compensation subnet, named CompenNeSt, which is designed with a siamese architecture to capture the photometric interactions between the projection surface and the projected images, and to use such information to compensate the geometrically corrected images.
  By concatenating WarpingNet with CompenNeSt, CompenNeSt++ accomplishes full projector compensation and is end-to-end trainable. 
  Third, to improve practicability, we propose a novel synthetic data-based pre-training strategy to significantly reduce the number of training images and training time.
  Moreover, we construct the first setup-independent full compensation benchmark to facilitate future studies. In thorough experiments, our method shows clear advantages over prior art with promising compensation quality and meanwhile being practically convenient.
\end{abstract}

}

\maketitle

\IEEEdisplaynontitleabstractindextext

% \fi
\IEEEpeerreviewmaketitle

\IEEEraisesectionheading{\section{Introduction}\label{sec:introduction}}
	
\begin{figure}[!t]
		\begin{center}
			\includegraphics[width=1\linewidth]{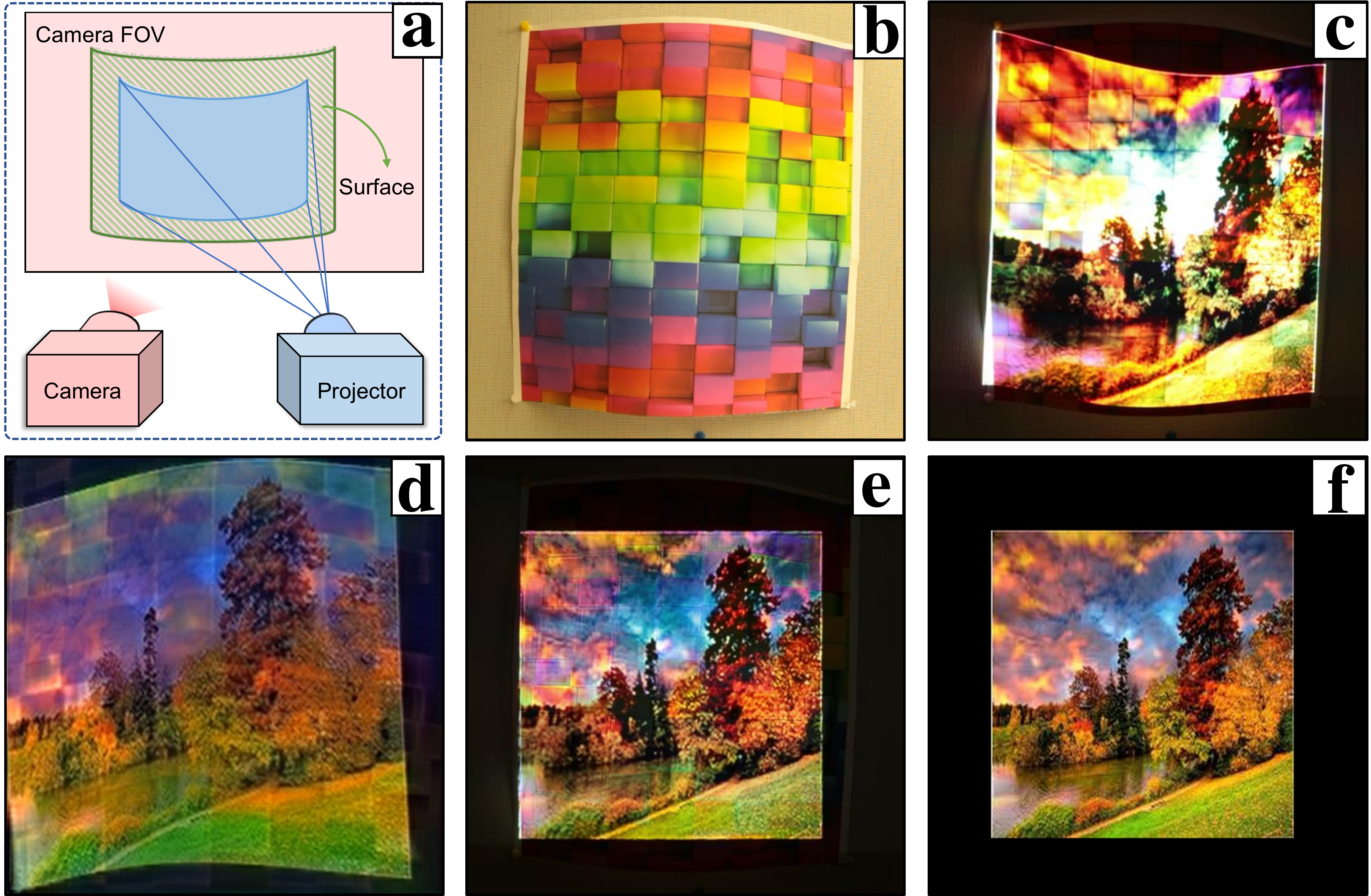}
			\caption{Full projector geometric correction and photometric compensation: \textbf{(a)} system setup with a nonplanar and textured surface \textbf{(b)}, \textbf{(c)} projection result without compensation, % showing obvious geometric and photometric distortion. 
				\textbf{(d)} fully compensated projector input image by our method, \textbf{(e)} camera-captured compensated projection result (\ie, (d) projected onto (b)), and \textbf{(f)} desired visual effect. Comparing (c) and (e) we see clearly improved geometry, color and texture details. }\label{fig:teaser}
		\end{center}
	\end{figure}

	\IEEEPARstart{W}{ith} the recent advance in projector technologies, projectors have been gaining increasing popularity with many applications~\cite{Geng2011,raskar2003ilamps, yoshida2003virtual,grossberg2004making, bimber2005embedded, aliaga2012fast,siegl2015real, siegl2017adaptive, narita2017dynamic, grundhofer2018recent, ueda2020illuminated, huang2018single, huang2020deltra}. However, nonplanar and textured projection surfaces are still challenging and often create bottlenecks for generalization of projector systems. Full projector geometric correction and photometric compensation \cite{raskar2001shader,bimber2005embedded, harville2006practical, siegl2015real, siegl2017adaptive, asayama2018fabricating} aims to address this issue by modifying a projector input image to compensate for the projection setup geometry \cite{raskar2001self, raskar2003ilamps, tardif2003multi, boroomand2016saliency, narita2017dynamic} and associated photometric environment \cite{yoshida2003virtual, ashdown2006robust, aliaga2012fast, grundhofer2015robust, huang2019compennet}. In the rest of the text, we call it \emph{full compensation} for conciseness. An example from our solution is illustrated in \autoref{fig:teaser}, where the compensated projection result (e) is clearly more visually pleasant than the uncompensated one in (c).

	A typical full projector compensation system consists of a projector-camera pair and a nonplanar textured projection surface placed at a fixed distance and orientation (\autoref{fig:teaser}(a)). Most existing methods work in two separate steps: (1) geometric surface modeling, \eg, via a sequence of structured light (SL) patterns \cite{Geng2011, moreno2012simple}, and (2) color and texture compensation on top of the geometrically corrected projection, via another sequence of sampling images. Once the camera captures all the projected sampling images, a composite function is fitted to map the input images to the captured images. This function (or its inverse) is then used to infer the compensated image for a new input image.
	
	Although relatively easy to implement, this two-step pipeline has four major issues:
	(1) Geometric correction is usually performed offline and typically requires \emph{additional} projected patterns (\eg, Gray-coded SL \cite{Geng2011, moreno2012simple}) that may be disturbed by the projection surface geometry (\eg, reflection, see \autoref{fig:compare_real}). It works well when the surface is textureless and the projector-camera system is photometrically calibrated. However, SL patterns decoding may be imperfect due to photometric disturbance \cite{Geng2011, shahpaski2017simultaneous, grundhofer2018recent} \eg, a textured surface or photometrically uncalibrated projector-camera settings or specular highlight, and thus may lead to an unreliable geometric correction. Then, a second step of photometric compensation on top of the erroneous geometric correction may aggravate the suboptimal solution.
	(2) Moreover, because the geometric correction is assumed independent of the photometric compensation, this two-step pipeline is \emph{non-differentiable} and thus inaccessible to derivative-based machine learning approaches.
	(3) Existing solutions (\eg, \cite{nayar2003projection,grossberg2004making, sajadi2010adict,grundhofer2015robust}) usually model the photometric compensation function explicitly, with various simplification assumptions that allow the parameters to be estimated from collected samples. These assumptions, such as context independence (\autoref{sec:photometric_compensation}), however, are often violated in practice. Furthermore, due to the extremely complex geometric and photometric processes involved in projector-camera systems, it is hard for traditional photometric compensation solutions to faithfully accomplish their task. 
	(4) When projector-camera system setup changes, \eg, replacing the projection surface, changing the projector-camera pose, and adjusting the environment light, the camera exposure or the projector brightness, existing methods must \emph{restart the projection-capturing-compensation process from scratch}, which is apparently impractical in real life applications. 
		
	In this paper, for the first time, an end-to-end full compensation solution is presented to address the above issues. We start by reformulating the compensation problem as a novel form (\ie, disentanglement of geometry and photometry) that can be learned online, as required by the compensation task in practice. 
	This formation allows us to develop a convolutional neural network (CNN), named \emph{CompenNeSt++}, to jointly solve both geometric correction and photometric compensation in a unified pipeline solely from the sampling images (\ie, without an additional step of structured light).

	By taking into consideration of both geometric and photometric ingredients in the compensation formulation, we carefully design CompenNeSt++ as composed of two subnets. 
	The first subnet is a novel cascaded coarse-to-fine geometric correction subnet, named \emph{WarpingNet} (\autoref{fig:flowchart}), which learns the sampling/warping grid and performs geometric correction; while the second subnet is a novel CNN named \emph{CompenNeSt} for photometric compensation. CompenNeSt consists of a siamese encoder and a decoder. Such an architecture captures rich multi-level interactions between the camera-captured projection surface image and the projector input image, and allows us to intuitively perform compensation by subtracting the surface features from the projector input image features through skip connections \cite{he2016deep}. Moreover, we use two low-level skip connections to carry the high frequency information to the penultimate and the output layers, allowing CompenNeSt to learn the residual features instead of inferring from scratch, thus improving the network convergence. 

	It is worth highlighting that the two subnets are concatenated directly (\autoref{fig:flowchart}), which makes CompenNeSt++ end-to-end trainable/differentiable, \ie, the loss gradients can back-propagate from CompenNeSt module to WarpingNet module, allowing joint optimization of geometry and photometry. 
	In addition, we propose multiple task-specific training strategies, such as projector field of view (FOV)-based WarpingNet weight initialization and projection-free CompenNeSt weight initialization to further improve model convergence.
	
	Another advantage of our CompenNeSt++ is practicability. When setup changes, traditional non-learning based methods have to rerun the compensation process from scratch, which is impractical in real life applications. On the contrary, we propose a Blender \cite{blender} rendered synthetic dataset and pre-train CompenNeSt on it, then CompenNeSt++ can be quickly fine-tuned to adapt to new setups. This strategy significantly reduces the number of training images (as few as 8 images) and training time (as short as 3 minutes) and is quite useful in practice. Moreover, during testing/inference, we simplify the WarpingNet to a single sampling grid and the CompenNeSt surface feature branch to biases, and thus further improve the running time efficiency of CompenNeSt++.

	Last but not least, an important issue addressed in this paper is the absence of evaluation benchmarks for projector compensation, due mainly to the fact that traditional evaluation is highly setup dependent. More specifically, to evaluate a compensation algorithm, theoretically, its experimental results need to be actually projected and captured and then quantitatively compared with ground truth. This process makes it impractical to provide a shared benchmark among different research groups. In this work, we tackle this issue by deriving a surrogate evaluation protocol that requires no actual projection of the algorithm output. As a result, this surrogate allows us to construct, for the first time, two sharable setup-independent compensation benchmarks, one for full compensation and another for partial photometric compensation. The proposed CompenNeSt++/CompenNeSt is evaluated on the two benchmarks that are carefully designed to cover various challenging factors. In the experiments, CompenNeSt++ demonstrates clear advantages compared with state-of-the-arts. 
	
	Our contributions can be summarized as follows:
	\begin{enumerate}
		\item For the first time, an end-to-end trainable solution is proposed for full compensation. Such a solution allows our system to effectively and explicitly capture the complex geometric and photometric interactions involved in the full compensation process.
		\item Compared with two-step methods, CompenNeSt++ not only is fully differentiable but also learns the geometric correction without an additional step of structured light and outperforms the traditional two-step methods.
		\item Two task-specific weight initializations and two network simplification techniques are proposed to further improve the convergence and running time efficiency of CompenNeSt++.
		\item For the first time, a synthetic data-based pre-training method is proposed to significantly improve the practical efficiency of our system.
		\item For the first time, a setup-independent full compensation benchmark and a partial photometric compensation benchmark are constructed, which is expected to facilitate future works in this direction.
	\end{enumerate}
	This paper builds upon preliminary conference papers CompenNet \cite{huang2019compennet} and CompenNet++ \cite{huang2019compennet++} and significantly extends them in various aspects. 	
	(1) We redesigned the \emph{photometric compensation subnet} as removing the surface from the projector input image in the feature space, based on which we propose a novel photometric compensation subnet named \emph{CompenNeSt} (\ie, the photometric part of CompenNeSt++, the additional ``S'' stands for siamese). Compared with our previous CompenNet \cite{huang2019compennet} (\ie, the photometric part of CompenNet++ \cite{huang2019compennet++}), CompenNeSt is designed with a siamese encoder to explicitly apply the same feature transformation to the surface image and the projector input image, using which we can perform photometric compensation by subtracting the surface pattern from the projector input image in the feature space. Moreover, minor tweaks on skip and transposed convolutional layers are shown to further improve model performance. We show in experimental comparisons that compared with CompenNet++ \cite{huang2019compennet++} (1,152,147 parameters), CompenNeSt++ architecture (833,227 parameters) not only reduces the number of learnable parameters by 27.7\% but also significantly improves full compensation accuracy; 
	(2) We perform in-depth ablation studies and show what features are learned by CompenNeSt++ and how the compensation is performed in the feature space. Such studies were not available in our previous papers.
	(3) We propose a novel pre-training strategy using Blender \cite{blender} rendered synthetic dataset, which greatly improves the practicability compared with the naive pre-training method of CompenNet/CompenNet++ \cite{huang2019compennet, huang2019compennet++} and it can significantly reduce the number of training images to 8 and training time to 3 minutes. It is worth noting that this paper focuses on static projector-camera setups and cannot compensate non-trivial setup changes without retraining, thus how to quickly adapt (instead of retraining from scratch) the system to new setups is particularly important.

	For the benefit of the society, the source code, evaluation benchmark and experimental results are publicly available at {\small\url{https://github.com/BingyaoHuang/CompenNeSt-plusplus}}.
 
	%%%%%%%%%%%%%%%%%%%%%%%%%%%%%%%%%%%%%%%%%%%%%%%%%%%%%%%%%%%%%%%%%%%%%%%%%%%%
	\section{Related Works}\label{sec:related_works}
	In theory, the projector compensation process is a very complicated nonlinear function involving the camera and the projector sensor radiometric responses \cite{nayar2003projection}, lens distortion/vignetting \cite{juang2007photometric}, perspective transformations \cite{huang2020flexible, zhang2000}, surface material reflectance \cite{phong1975illumination,huang2020deltra}, defocus \cite{zhang2006projection, yang2016practical,kageyama2020prodebnet} and inter-reflection \cite{takeda2016inter}. A great amount of effort has been dedicated to designing practical and accurate compensation models, which can be roughly categorized into two types: full compensation \cite{raskar2001shader,bimber2005embedded, harville2006practical, wetzstein2007radiometric, siegl2015real, siegl2017adaptive, shahpaski2017simultaneous} and partial ones \cite{nayar2003projection, grossberg2004making, sajadi2010adict,  grundhofer2015robust,ashdown2006robust,aliaga2012fast, takeda2016inter, li2018practical, huang2019compennet}. 
	
	\subsection{Full compensation methods}
	Full compensation methods perform both geometric correction and photometric compensation. The pioneer work by Raskar \etal~\cite{raskar2001shader} creates projection mapping animations on nonplanar colored objects with two projectors. Despite compensating both geometry and photometry, manual registrations using known markers are required.
	Wetzstein \etal~\cite{wetzstein2007radiometric} employ a full light transport matrix for full compensation. Despite obtaining accurate global illumination and geometry, it requires an additional radiometric calibration step and capturing and inverting the full light transport matrix is relatively time consuming.
	Harville \etal~\cite{harville2006practical} propose a full multi-projector compensation method for a white curved screen. The projector-camera pixel correspondences are obtained via 8-12 SL images. Despite being effective to blend multiple projector's colors, this method assumes a textureless projection surface.
	
	Recently, Siegl \etal~\cite{siegl2015real, siegl2017adaptive} perform full compensation on nonplanar Lambertian surfaces for dynamic real-time projection mapping. Similar to \cite{harville2006practical}, they assume the target objects are white and textureless. Asayama \etal~\cite{asayama2018fabricating} attach visual markers to nonplanar textured surfaces for real-time object pose tracking. To remove the disturbance of the markers, photometric compensation is applied to hide the markers from the viewer, and additional IR cameras/emitters are required accordingly. Shahpaski \etal~\cite{shahpaski2017simultaneous} embed color squares in the projected checkerboard pattern to calibrate both the geometry and the gamma function. Although only two shots are required, this method needs a pre-calibrated camera and another planar printed checkerboard target. Moreover, it only performs a uniform gamma compensation without compensating the surface, and thus may not work well on nonplanar textured surfaces. 

	\subsection{Partial compensation methods}
	Compared to full compensation methods, partial compensation ones typically perform either geometric correction \cite{raskar2001self, raskar2003ilamps, tardif2003multi, boroomand2016saliency, narita2017dynamic} or photometric compensation \cite{yoshida2003virtual, ashdown2006robust, aliaga2012fast, grundhofer2015robust, huang2019compennet}. Due to the strong mutual-dependence between geometric correction and photometric compensation and to avoid propagated errors from the other part, these methods assume the other part is already performed as a prerequisite. However, this two-step pipeline is non-differentiable and may be subject to suboptimal solutions. 
	
	\subsubsection{Geometric correction} 
	Without using specialized hardware, such as a coaxial projector-camera pair \cite{fujii2005projector}, projector-camera image pairs' geometric mapping needs to be estimated using methods such as structured light (SL) \cite{raskar2001self,raskar2003ilamps, tardif2003multi, boroomand2016saliency}, markers \cite{narita2017dynamic} or homographies \cite{huang2019compennet}. Raskar \etal~\cite{raskar2003ilamps} propose a conformal texture mapping method to geometrically register multiple projectors for nonplanar surface projections, using SL and a calibrated camera. Tardif \etal~\cite{tardif2003multi} achieve similar results without calibrating the projector-camera pair.  The geometrically corrected image is generated by SL inverse mapping. Similarly, Boroomand \etal~\cite{boroomand2016saliency} propose a saliency-guided SL geometric correction method.
	Narita \etal~\cite{narita2017dynamic} use IR ink printed fiducial markers and a high-frame-rate camera for dynamic non-rigid surface projection mapping, which requires additional devices as \cite{asayama2018fabricating}.
	
	\subsubsection{Photometric compensation}\label{sec:photometric_compensation}
	These methods assume that the projector-camera image pairs are registered as a prerequisite and can be roughly categorized into two types: context-independent \cite{nayar2003projection, grossberg2004making, sajadi2010adict,  grundhofer2015robust} and context-aware ones \cite{ashdown2006robust,aliaga2012fast, takeda2016inter, li2018practical, huang2019compennet}, where context-aware ones typically assume projector-camera pixels one-to-one mapping while context-aware ones also consider neighborhood/global information. A detailed review can be found in \cite{grundhofer2018recent}.

	\noindent\textbf{Context-independent methods} typically assume that there is an approximate one-to-one mapping between the projector and camera image pixels, \ie, a camera pixel only depends on its corresponding projector pixel and the surface patch illuminated by that projector pixel. Namely, each pixel is roughly independent of its neighborhood context.
	The pioneer work by Nayar \etal~\cite{nayar2003projection} proposes a linear model that maps a projector ray brightness to camera detected irradiance with a 3$\times$3 color mixing matrix. Grossberg \etal~\cite{grossberg2004making} improve Nayar's work and model the environment lighting by adding a 3$\times$1 vector to the camera-captured irradiance. However, an additional step of the uniform radiometric responses calibration is required for the linearity to hold. Moreover, as pointed out by Juang \etal~\cite{juang2007photometric}, even with radiometric calibration, the assumption of uniform radiometric response may be violated. 
	
	To address this issue, some studies absorb the nonlinear radiometric responses in the compensation function, \eg, Sajadi \etal~\cite{sajadi2010adict} fit a smooth higher-dimensional B{\'e}zier patches-based model with 9\textsuperscript{3}=729 sampling images. Grundh{\"o}fer and Iwai \cite{grundhofer2015robust} propose a thin plate spline (TPS)-based method and reduce the number of sampling images to 5\textsuperscript{3}=125 and further deal with clipping errors and image smoothness with a global optimization step.
	Other than optimizing the image colors numerically, some methods specifically focus on human perceptual properties, \eg, Huang \etal~\cite{huang2017radiometric} generate visually pleasing projections by exploring human visual system's chromatic adaptation and perceptual anchoring property. Also, clipping artifacts due to camera/projector sensor limitation are minimized using gamut scaling.

	Despite largely simplifying the compensation problem, the context-independent assumption is usually violated in practice, due to many factors such as perspective projection, lens distortion, defocus and surface inter-reflection \cite{zhang2006projection, wetzstein2007radiometric, yang2016practical,takeda2016inter}. As a result, a projector pixel can illuminate multiple surface patches and a patch can also be illuminated by the inter-reflections of its surrounding patches, and a camera pixel is also determined by rays reflected by multiple patches. 
	
	\noindent\textbf{Context-aware methods} compensate a pixel by considering information from neighborhood context. Grundh{\"o}fer and Bimber \cite{grundhofer2008real} tackle visual artifacts and enhance brightness and contrast by analyzing the projection surface and input image prior. Li \etal~\cite{li2018practical} reduce the number of sampling images to at least two by sparse sampling and linear interpolation. Multidimensional reflectance vectors are extracted as color transfer function control points.
	Besides computing an offline compensation model, Aliaga \etal~\cite{aliaga2012fast} introduce a run time linear scaling operation to optimize multiple projector compensations.  Takeda \etal~\cite{takeda2016inter} propose an inter-reflection compensation method using an ultraviolet LED array.
	
	Context-aware methods generally improve over context-independent methods by integrating more information. However, it is extremely hard to model or approximate the ideal compensation process due to complex interactions between the global illuminations, the projection surface and the input image. Moreover, most existing works focus on reducing pixel-wise color errors only rather than jointly improving the color fidelity and structural similarity \cite{wang2004image}.

	\subsection{Our method}
	To the best of our knowledge, there exists no previous end-to-end trainable method that performs simultaneous full projector geometric correction and photometric compensation. Our method 
	is inspired by the successes of recently proposed deep learning-based image-to-image translation, such as Pix2pix \cite{isola2017image}, CycleGAN \cite{zhu2017unpaired}, neural style transfer \cite{johnson2016perceptual, gatys2016image, huang2017arbitrary}, image super-resolution \cite{dong2014learning, kim2016accurate, ledig2017photo, wang2018sftgan} and image colorization \cite{zhang2016colorful, iizuka2016let, deshpande2017learning}. That said, as the first deep learning-based full compensation algorithm, our method is very different from these studies and has its own special physical domain knowledge. For example, unlike the CNN models above that can be trained once and for all, the projector compensation model needs to be quickly retrained if the system setup changes. However, in practice, capturing training images and training the model are both time consuming. In addition, data augmentation techniques such as random cropping, affine transformation and color jitter are not available for our task, because each camera pixel is strongly coupled with a neighborhood of its corresponding projector pixel and the projection surface patch illuminated by those pixels. Furthermore, general image-to-image translation models do not consider the physical domain knowledge of projector compensation task, e.g., they do not 	explicitly formulate the complex geometric transformations and spectral interactions between the global lighting, the projector backlight and the projection surface. In fact, the advantage of the proposed method over the classical Pix2pix \cite{isola2017image} algorithm is clear in our evaluations.

	It is worth noting that there are some efficient \emph{partial} compensation methods, \eg, Kurth \etal \cite{kurth2018auto} perform \emph{geometric} calibrations for multi-projectors in less than 2 minutes, but it requires a precise mesh of the target object, and the object should also be mostly white and diffuse. Li \etal \cite{li2018practical} present a real-time \emph{photometric} compensation method with only two sampling images. However, due to the small size of sampling dots, this method may be sensitive to projector defocus and high frequency surface textures. A simple linear interpolation using those unreliable samples may reduce the compensation quality.
	
	In summary, belonging to the full compensation regime, our CompenNeSt++ is the first to jointly learn geometric correction and photometric compensation in an end-to-end framework. The advantage of the proposed CompenNeSt++ over both traditional and deep learning-based two-step methods, is clearly demonstrated quantitatively and qualitatively.

	\begin{figure*}[!t]
		\begin{center}
		\includegraphics[width=1\linewidth]{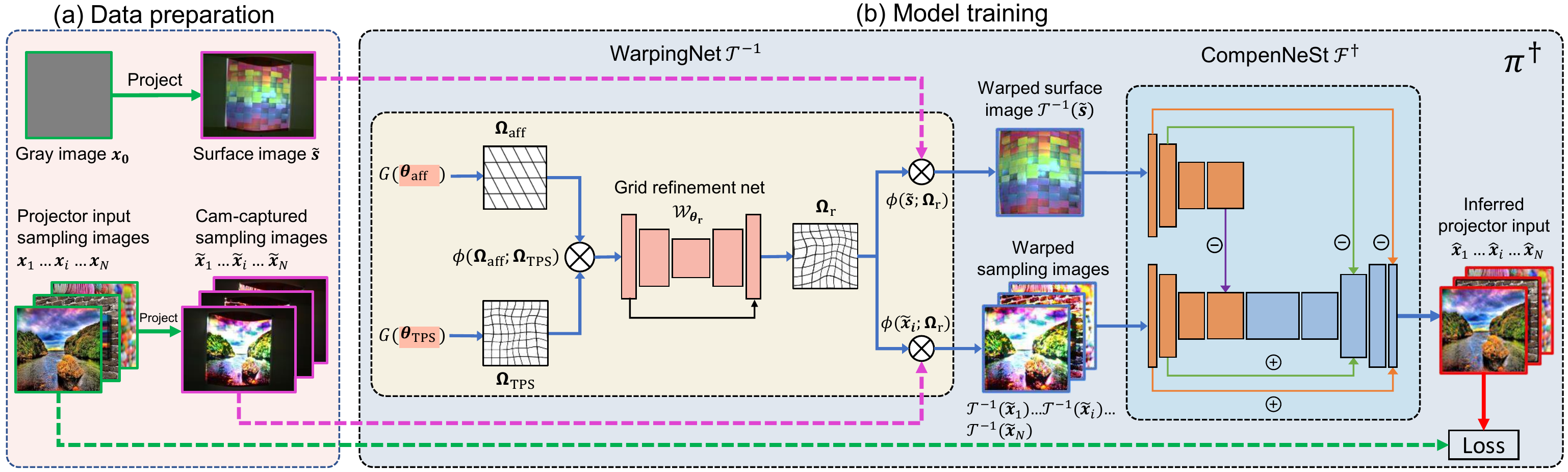}
			\caption{Architecture of the proposed CompenNeSt++ and its
				training in two major steps. \textbf{(a)} Project and capture a surface image and a set of sampling images. \textbf{(b)} CompenNeSt++, \ie, $\pi^{\dagger}_{\vec{\theta}}$, is trained using the data prepared in (a). The projector input images are outlined in green, the camera-captured images are outlined in purple, the intermediate results (\eg, warped images) are outlined in blue and the network output images are outlined in red. The dashed lines indicate network input training data.
				\textbf{WarpingNet} ({\small$ \T^{-1} $}, yellow block) warps the camera-captured images $ \tilde{\s} $ and $ \tilde{\x} $ to the projector canonical frontal view using a cascaded coarse-to-fine structure, where the pink modules are learnable parameters. Operator $ \otimes $ denotes a bilinear interpolator, \ie, $ \phi(\cdot; \cdot) $. The grid refinement network {\small$\mathcal{W}_{\vec{\theta}\textsubscript{r}}$} consists of a UNet-like \cite{ronneberger2015u} structure, it generates a refined sampling grid that is used to sample (warp) the input images.
				\textbf{CompenNeSt} ({\small$ \F^{\dagger} $}, light blue block) consists of a siamese encoder (orange modules share weights) and a decoder (blue modules). Best viewed in color.
			}
			\label{fig:flowchart} 
			\end{center}
	\end{figure*}

	%%%%%%%%%%%%%%%%%%%%%%%%%%%%%%%%%%%%%%%%%%%%%%%%%%%%%%%%%%%%%%%%%%%%%%%%%%%%
	\section{Deep Full Projector Compensation}\label{sec:problem}
	\subsection{Problem formulation}\label{subsec:problem_formulation}
	
	Our full projector compensation system consists of an uncalibrated projector-camera pair and a nonplanar textured projection surface placed at a fixed distance and orientation (\autoref{fig:teaser}(a)). Denote the projector input image as $\x$, the composite geometric projection and photometric transfer function as $ \pi_p $ and the projector geometric and photometric parameters as $ \vec{p} $. Then, the projected radiance can be denoted as $\pi_p(\x, \vec{p})$. Let the composite surface reflectance, geometry and pose be $\s$, the surface reflection function be $\pi_s$,  the global lighting be $\vec{g}$, camera's composite capturing function be $\pi_c$, and its composite parameters be $ \vec{c} $. Finally, the camera-captured image $\tilde{\x}$ is given by\footnote{We use `tilde' ($\tilde{\x}$) to indicate a camera-captured image, see \autoref{fig:flowchart}(a).}:
	\begin{equation}\label{eq:ref}
	\tilde{\x} = \pi_c\big(\pi_s\big(\pi_p( \x , \vec{p} ), \vec{g}, \s\big), \vec{c}\big)
	\end{equation}	
	Note that the composite geometric and photometric process in \autoref{eq:ref} is very complex and obviously has no closed form solution. Instead, we find that $\vec{p}$ and $\vec{c}$ are constant once the setup is fixed, thus, we disentangle the geometric and photometric transformations and absorb $\vec{p}$ and $\vec{c}$ in two functions: {\small$\T:\mathbb{R}^{H_1\times W_1\times 3}\mapsto \mathbb{R}^{H_2\times W_2\times 3}$} that geometrically warps a projector input image to camera-captured image; and {\small$\F:\mathbb{R}^{H_1\times W_1\times3}\mapsto \mathbb{R}^{H_1\times W_1\times3} $} that photometrically transforms a projector input image to an uncompensated camera-captured image (aligned with the projector canonical frontal view). Thus, \autoref{eq:ref} can be reformulated as:
	\begin{equation}\label{eq:forward}
	\tilde{\x} = \T(\F(\x; \vec{g}, \s))
	\end{equation}
	
	Full projector compensation aims to find a projector input image $ \x^{*} $, named \emph{compensation image} of $ \x $, such that the viewer-perceived projection result is the same as the ideal desired viewer-perceived image $\x'$, \ie,
	\begin{equation}\label{eq:cmp_cam}
	\T(\F(\x^{*}; \vec{g}, \s))  = \x'
	\end{equation}
	where $\x'$ is an affine transformed $\x$ to match the optimal displayable area (\autoref{fig:fov_mask} and \autoref{fig:testing}).
	Thus the  compensation image $ \x^{*} $ in \autoref{eq:cmp_cam} can be solved by:
	\begin{equation}\label{eq:x*}
	\x^{*} = \F^{\dagger}(\T^{-1}(\x'); \vec{g}, \s).
	\end{equation}
	
	In practice it is hard to measure $ \vec{g} $ and $ \vec{s} $ directly. For this reason, we implicitly capture them using a camera-captured surface image $\tilde{\s} $ under the global lighting and the projector backlight:
	\begin{equation}\label{eq:g_s}
	\tilde{\s} = \T(\F(\x_0; \vec{g}, \s)),
	\end{equation}
	where $\x_0$ is set to a plain gray image to provide some illumination.
	
	It is worth noting that other than the surface patches illuminated by the projector, the rest part of the surface outside the projector FOV does not provide useful information for compensation (\autoref{fig:flowchart}(a) black regions of $\tilde{\s}$), thus $\tilde{\s}$ in \autoref{eq:g_s} can be approximated by a subregion of the camera-captured image $ \T^{-1}(\tilde{\s})$ (\autoref{fig:flowchart}(b)).	Substituting $\vec{g}$ and $\s$ in \autoref{eq:x*} with $\T^{-1}(\tilde{\s})$ , we have the compensation problem as
	\begin{equation}\label{eq:compensation}
	\x^{*} = \F^{\dagger}\big(\T^{-1}(\x'); \T^{-1}(\tilde{\s})\big),
	\end{equation}
	where $ \F^{\dagger} $ is the pseudo-inverse of $ \F $ and $ \T^{-1} $ is the inverse geometric transformation of $ \T $. 
	
	\begin{figure*}[!t]
		\begin{center}
			\includegraphics[width=1.0\linewidth]{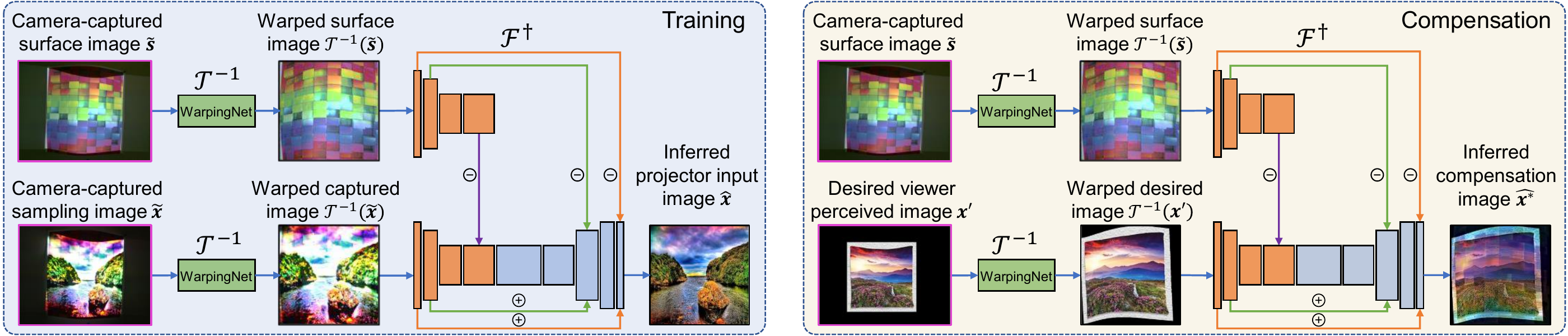}
			\caption{ We train CompenNeSt++ using camera-captured and projector input image pairs like $ (\tilde{\x}, \x) $ instead of desired viewer-perceived and compensation image pairs like $ (\x', \x^*) $, obviating the need for the ground truth compensation image $\x^*$, because learning the backward mapping from the camera-captured uncompensated image to the projector input image (left: $ \tilde{\x} \mapsto  \x $) (\autoref{eq:training}) is the same as learning the backward mapping from the desired viewer-perceived image to the projector compensation image (right: $ \x' \mapsto  \x^*$) (\autoref{eq:compensation}).
				}\label{fig:surrogate_train}
		\end{center}
	\end{figure*}
	
	\subsection{Learning-based formulation}
	Obviously, \autoref{eq:compensation} has no closed form solution and thus we model $\F^{\dagger}$ and $ \T^{-1} $ jointly with a deep neural network named \emph{CompenNeSt++} and learn it using image pairs like $ (\x^{*}, \x') $ and a camera-captured surface image $ \tilde{\s} $. A key requirement for this solution is the availability of training data, however it is very difficult to obtain the ground truth of the compensation image $ \x^{*} $. Fortunately, by investigating \autoref{eq:forward} and \autoref{eq:g_s} we find that:
	\begin{equation}\label{eq:training}
	\tilde{\x} = \T\big(\F(\x; \tilde{\s})\big) \ \ \Rightarrow \ \ \x = \F^{\dagger}\big(\T^{-1}(\tilde{\x}); \T^{-1}(\tilde{\s})\big)
	\end{equation}
	The equation above follows the same physical process as \autoref{eq:compensation}, indicating that we can train CompenNeSt++ over sampled image pairs like $(\tilde{\x}, \x)$ and a surface image $\tilde{\s}$\footnote{$ \tilde{\vec{s}} $ is fixed as long as the the setup is unchanged, thus only one $ \tilde{\vec{s}} $ is needed for training and testing/inference, see \autoref{fig:flowchart}(a).}, which can be easily obtained (\autoref{fig:surrogate_train} left). 
	
	Another advantage of learning CompenNeSt++ using \autoref{eq:training} instead of \autoref{eq:compensation} is that we can construct a sharable setup-independent benchmark for model evaluation and comparison without actual projections or captures, see \autoref{sec:benchmark}.

	In the rest of the paper, we abuse the notation {\small$ \pi^{\dagger}_{\vec{\theta}}(\cdot, \cdot) \equiv \F^{\dagger}\big(\T^{-1}(\cdot); \T^{-1}(\cdot)\big)$} for conciseness, where $ \vec{\theta}= \{\vec{\theta}_\F ,\vec{\theta}_\T\} $ are CompenNeSt++'s learnable parameters.
	By using \autoref{eq:training}, we can capture a set of $N$ training pairs, denoted as $\mathcal{X}=\{(\tilde{\x}_i, \x_i)\}_{i=1}^N $. Then, with a proper image reconstruction loss $ \mathcal{L}$, CompenNeSt++ can be learned by
	\begin{equation}\label{eq:objective}
	\vec{\theta} = \argmin_{\vec{\theta}'}\sum_i\mathcal{L}\big(\vec{\hat{x}}_i=\pi^{\dagger}_{\vec{\theta}'}(\tilde{\x}_i; \tilde{\s}), \ \x_i\big)
	\end{equation}
	where $ \vec{\hat{x}} $ is the compensation of $\tilde{\x}$ (not $\x$, see \autoref{fig:flowchart} network output). Unlike previous methods \cite{grundhofer2015robust, sajadi2010adict} that use simple pixel-wise $ \ell_2 $ losses, we use the loss function below to jointly optimize the color fidelity (pixel-wise $ \ell_1 $) and structural similarity (SSIM) \cite{wang2004image}. 
	\begin{equation}\label{eq:loss}
	\mathcal{L}  = \mathcal{L}_{\ell_1} + \mathcal{L}_{\text{SSIM}}
	\end{equation}
	The advantages of this loss function are shown in \cite{zhao2017loss} and experimental comparisons in \autoref{subsec:loss}.
	
	\subsection{Network design}\label{subsec:network_design}
	Based on the above formulation, our CompenNeSt++ is designed with two subnets, a \emph{WarpingNet} {\small$ \T^{-1} $} that corrects the geometric distortions and warps camera-captured uncompensated images to the projector canonical frontal view; and a \emph{CompenNeSt} {\small$ \F^{\dagger} $} that photometrically compensates warped images. The network architecture is shown in \autoref{fig:flowchart}.  For compactness, we move the detailed parameters of CompenNeSt++ to the \href{https://vision.cs.stonybrook.edu/~bingyao/pub/CompenNeSt_supp}{supplementary material}.
	
	\subsubsection{WarpingNet $\T^{-1}$ }\label{subsec:warpingnet}
	Note that directly estimating nonparametric geometric corrections is difficult and computationally expensive \cite{shen2019networks, revaud2016deepmatching}. Instead, we model the geometric correction as a cascaded coarse-to-fine process, as inspired by the work in~\cite{jaderberg2015spatial,rocco2017convolutional}. As shown in \autoref{fig:flowchart}, WarpingNet consists of three learnable modules (\ie, $ \vec{\theta}\textsubscript{aff}$, $\vec{\theta}$\textsubscript{TPS} and {\small$\mathcal{W}_{\vec{\theta}\textsubscript{r}}$}), a grid generation function $ G $, a bilinear interpolation-based image sampler $ \phi $,  and three generated sampling grids ranked in order of increasing granularity: {\small
		$\vec{\Omega}\textsubscript{aff}=G(\vec{\theta}\textsubscript{aff}),
		\vec{\Omega}\textsubscript{TPS}=G(\vec{\theta}\textsubscript{TPS}), 
		\vec{\Omega}\textsubscript{r}=\mathcal{W}_{\vec{\theta}\textsubscript{r}}(\vec{\Omega}\textsubscript{TPS})$}.

	Specifically, 	$ \vec{\theta}\textsubscript{aff}$ is a 2$\times$3 learnable affine matrix and it roughly warps the input image $ \tilde{\x} $ to the projector canonical frontal view. Similarly, $ \vec{\theta}\textsubscript{TPS}$ contains (6$\times$6+2)$\times$2 =76 learnable thin plate spline (TPS) \cite{donato2002approximate} parameters and it further nonlinearly warps the rough affine-transformed image $ \phi(\tilde{\x}; \vec{\Omega}\textsubscript{aff}) $ to better match the exact projector canonical frontal view. Unlike \cite{jaderberg2015spatial, rocco2017convolutional}, $ \vec{\theta}\textsubscript{aff}$ and $ \vec{\theta}\textsubscript{TPS}$ are directly learned without using a regression network, which is more efficient and accurate in our case.
	
	Although TPS can approximate nonlinear smooth geometric transformations, its accuracy depends on the number of control points and the spline assumptions. Thus, it may not precisely model geometric deformations involved in the projector-camera imaging process. To address this issue, we design a grid refinement CNN, \ie, $ \mathcal{W}_{\vec{\theta}\textsubscript{r}} $ to refine the TPS sampling grid. Basically, this module learns a fine-grained displacement for each 2D coordinate in the TPS sampling grid by a residual connection \cite{he2016deep}, giving the refined sampling grid $ \vec{\Omega}\textsubscript{r}$ higher precisions. The advantages of our CompenNeSt++ over a degraded CompenNeSt++ without grid refinement net (named CompenNeSt++ w/o refine) are evidenced in \autoref{tab:compare}.
	
	\begin{figure*}[t]
		\begin{center}
			\includegraphics[width=1.0\linewidth]{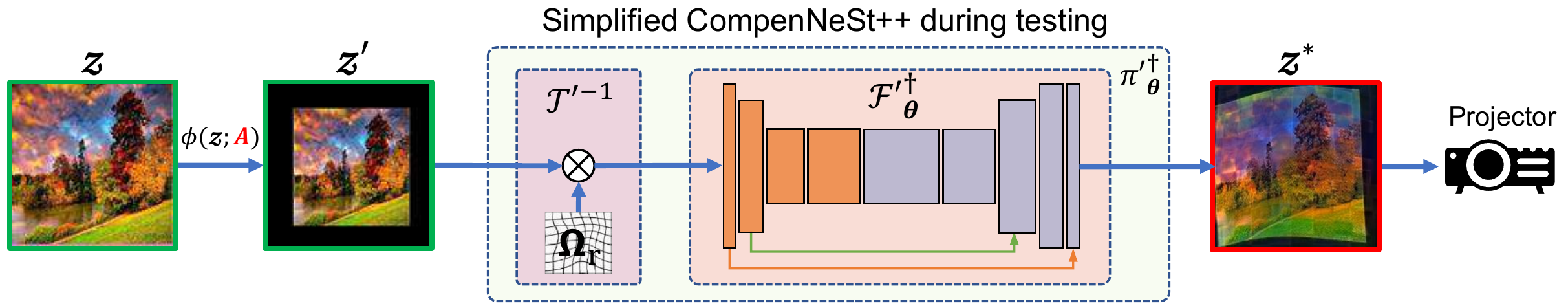}
			\caption{CompenNeSt++ in testing/inference phase. Due to the novel structure and sampling strategy, WarpingNet $ \T^{-1} $ can be simplified to a single sampling grid $\vec{\Omega}\textsubscript{r}$ and an image interpolator $ \otimes $. Moreover, CompenNeSt surface features can be integrated into the backbone as biases. Both techniques improve computational and memory efficiency during testing/inference with no performance drop. Finally, the model inferred compensation image $ \vec{z}^{*}  $ is both geometrically and photometrically compensated, such that after projection it cancels the geometric and photometric distortions and produces an image that is close to $\vec{z}'$, \ie, \autoref{fig:teaser}(e). More camera-captured compensation results are shown in \autoref{fig:compare_real} and \href{https://vision.cs.stonybrook.edu/~bingyao/pub/CompenNeSt_supp}{supplementary material}.
			}\label{fig:testing}
		\end{center}
	\end{figure*}

	Besides the novel cascaded coarse-to-fine structure with a grid refinement network, we propose a novel sampling strategy that improves WarpingNet efficiency and accuracy. Intuitively, the cascaded coarse-to-fine sampling method should sequentially sample the input $ \tilde{\x} $ by
	\begin{equation}\label{eq:img_sampling}
	\T^{-1}(\tilde{\x}) = \phi\big(\phi(\phi(\tilde{\x}; \vec{\Omega}\textsubscript{aff}); \vec{\Omega}\textsubscript{TPS});\vec{\Omega}\textsubscript{r} \!\!=\!\!\mathcal{W}_{\vec{\theta}\textsubscript{r}}(\vec{\Omega}\textsubscript{TPS})\big)
	\end{equation}	
	However, the three bilinear interpolations ($ \phi $) above not only are computationally inefficient but also produce a blurred image. Instead, we perform the sampling in the 2D coordinate space, \ie, let the finer TPS grid sample the coarser affine grid followed by a refinement using $ \mathcal{W}_{\vec{\theta}\textsubscript{r}} $,  as shown in \autoref{fig:flowchart}(b). Thus, the new warped image is given by:
	\begin{equation}\label{eq:grid_sampling}
	\T^{-1}(\tilde{\x}) =	\phi\big(\tilde{\x}; \mathcal{W}_{\vec{\theta}\textsubscript{r}}(\phi(\vec{\Omega}\textsubscript{aff}; \vec{\Omega}\textsubscript{TPS})\big)
	\end{equation}
	This strategy brings two benefits: (1) only two sampling operations are required and thus it is more efficient; and (2) since the image sampling is only performed once on $ \tilde{\x} $, the warped image is sharper compared with using \autoref{eq:img_sampling}.

	Another novelty of WarpingNet is network simplification owing to the sampling strategy above. During testing, WarpingNet is simplified essentially to a single sampling grid $\vec{\Omega}\textsubscript{r} $, and thus the geometric correction becomes a single bilinear interpolation {\small $ \T^{-1}(\tilde{\x}) =  \phi(\tilde{\x};\vec{\Omega}\textsubscript{r}) $}. This strategy allows us to perform geometric corrections without running grid generation or refinement network, bringing improved testing/inference efficiency (see \autoref{fig:testing}).  
	
	Finally, to improve training convergence and robustness we introduce carefully designed WarpingNet weights initialization techniques in \autoref{subsec:domain_knowledge}.

	\subsubsection{CompenNeSt $\F^{\dagger}$}\label{subsec:compennet}
	CompenNeSt consists of a siamese encoder and a decoder. During training (\autoref{fig:flowchart}(b) and \autoref{fig:surrogate_train} left), it takes two WarpingNet transformed camera-captured images as inputs, \ie, a warped surface image {\small$ \T^{-1}(\tilde{\s}) $} and a warped sampling image {\small$ \T^{-1}(\tilde{\x}) $} and outputs the inferred projector-input image {\small$ \hat{\x}$}. Both two inputs and the output are 256$\times$256$\times$3 RGB images. 
	Firstly, $ \tilde{\s} $ and $ \tilde{\x} $ are fed to the siamese encoder to downsample and to extract multi-level feature maps. Note that in \autoref{fig:flowchart} we give the two encoder branches the same orange color to emphasize that they share weights. Then, the surface's multi-level and multi-scale feature maps 
	are \emph{subtracted} from the feature maps of camera-captured image $ \tilde{\x} $. This design is inspired by the observation that compensation is analogous to removing/cancelling the surface ({\small$ \T^{-1}(\tilde{\s}) $}) disturbance/patterns/features from the input images ({\small$ \x'$ or $ \tilde{\x}$}). See how to interpret this physical domain knowledge in \autoref{subsec:ablation_act}. 
	
	To improve convergence, we pass low-level interaction information (\ie, feature maps from the first two layers) to high-level feature maps through skip convolutional layers \cite{he2016deep}. This design is based on the observation that the output compensated image should look like the projector input image on \emph{structure}, thus passing the low-level features (\ie, high frequency structural information) to the output layer allows the network to learn residuals on top of a good initial guess thus obviating the need of inferring from scratch.

	However, even with the above structure we find it difficult to jointly learn geometric and photometric processes without a proper model initialization, and the output images may become plain gray. To address this issue, we incorporate rich task-specific domain knowledge to weights initialization and training strategies below.
		
	\subsection{Task-specific domain knowledge and constraints}\label{subsec:domain_knowledge}
	To improve model convergence and robustness, we leverage rich task-specific domain knowledge of projector-camera systems to initialize and train CompenNeSt++.

	\subsubsection{Projector FOV mask} 
	According to \autoref{eq:training}, full projector compensation's region of interest is the projector FOV, \ie, \autoref{fig:fov_mask} illuminated regions. Thus we can compute a projector FOV mask by automatically thresholding the camera-captured surface images with Otsu's method \cite{otsu1979threshold} followed by some morphological operations (\autoref{fig:fov_mask}). This mask brings threefold benefits: \textbf{(1)} masking out the pixels outside of the projector FOV improves training stability and efficiency because the image reconstruction loss (\autoref{eq:loss}) increases significantly when black regions are mis-registered to the ground truth $ \x $, forcing the WarpingNet to quickly infer a plausible warping grid; \textbf{(2)} the projector FOV mask is the key to initialize WarpingNet affine weights $\vec{\theta}$\textsubscript{aff} in \autoref{subsec:warpingnet_init} and \textbf{(3)} to find the optimal displayable area in \autoref{subsec:pipeline}.
	
	\subsubsection{WarpingNet weights initialization}\label{subsec:warpingnet_init}
	We further improve the training efficiency by providing a task-specific prior, \eg, the coarse affine warping branch in WarpingNet (\autoref{fig:flowchart}(b)) aims to transform the input image $ \tilde{\x} $ to the projector canonical frontal view, as mentioned in \autoref{subsec:warpingnet}. Thus, the affine parameters $\vec{\theta}$\textsubscript{aff} can be initialized such that the projector FOV mask's bounding rectangle (\autoref{fig:fov_mask} green rectangle) is stretched to fill the warped image. Then, to avoid implausible large displacement from a default random initialization, $\vec{\theta}$\textsubscript{TPS} and grid refinement net $\mathcal{W}_{\vec{\theta}\textsubscript{r}}$ are initialized with small random numbers at a scale of $10^{-4}$, such that they generate identity mappings. These task specific initialization techniques provide a relatively good starting point, allowing CompenNeSt++ to converge stably and efficiently.
	
	\subsubsection{CompenNeSt weights initialization} \label{subsec:compennest_init}
	In our end-to-end full compensation pipeline, despite with the training techniques of WarpingNet above, joint training WarpingNet and CompenNeSt may still subject to suboptimal solutions, \eg, the output images become plain gray. Similar to WarpingNet weights initialization, we introduce some photometric prior knowledge to improve CompenNeSt stability and efficiency. 

	Since our CompenNeSt is has an encoder-decoder-like structure, the weights can be initialized by setting the input \emph{surface} image $ \T^{-1}(\tilde{\s}) $ to zero and training the model in an antoencoder way, \ie, reconstructing the input camera-captured sampling image $ \T^{-1}(\tilde{\x}) $. We further simplify the input camera-captured sampling image $ \T^{-1}(\tilde{\x}) $ to a projector input image $ \x $ to avoid actual projection. The training objective function is given in \autoref{eq:init}. 
	
	\begin{equation}\label{eq:init}
 		\vec{\theta}_\F = \argmin_{\vec{\theta}_\F'}\sum_i\mathcal{L}\big(\vec{\hat{x}}_i=\F^{\dagger}_{\vec{\theta}_\F'}({\x}_i; \vec{0}), \ \x_i\big)
	\end{equation}
	We train the model on 500 textured sampling images for 2,000 iterations. Then in practice, CompenNeSt can be initialized by loading the saved weights.

	\subsection{Practical compensation using a pre-trained model} \label{subsec:pre-train}
	Projector compensation requires re-calibration once the setup changes, \eg, replacing the projection surface, changing the projector-camera pose, and adjusting the environment light, the camera exposure or the projector brightness. Previous approaches must \emph{rerun the projection-capturing-compensation process from scratch}, which is apparently impractical in real life applications. On the contrary, owing to the end-to-end trainable architecture, our CompenNeSt can be pre-trained and then fine-tuned on new setups using much fewer images and less training time.

	Firstly, we render 100 setups with different projector-camera-surface poses, materials, exposures and lightings in Blender \cite{blender} (see \autoref{subsec:synthetic_data}). Then, we initialize CompenNeSt using the technique in \autoref{subsec:compennest_init} and trained CompenNeSt $ \F^{\dagger} $ alone on the synthetic dataset for 20k iterations. 
	Finally, for a new setup, {\small $\vec{\theta}_{\F}$} is initialized by loading the saved weights. This pre-training approach provides a powerful initialization without any actual projection/capture and the pre-training is performed only once and independent of setups. Compared with previous approaches, our pre-training method not only greatly reduces the number of training images and training time (see \autoref{subsec:exp_pre-train}), but also may facilitate future learning-based projector compensation methods via the collected dataset.

	\begin{figure}[!t]
		\begin{center}
			\includegraphics[width=1\linewidth]{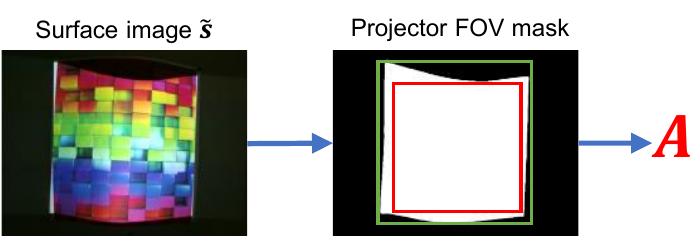}
			\caption{Projector FOV mask, bounding rectangle (green) and optimal displayable area (red). The optimal displayable area is defined as the maximum inscribed rectangle (keep aspect ratio) \cite{raskar2003ilamps}. The affine matrix $ \vec{A} $ is estimated given the displayable area and the projector input image size.}
			\label{fig:fov_mask}
		\end{center}
	\end{figure}
	
	\subsection{Network Simplification}\label{subsec:cnn_testing}
	To further improve model running time efficiency during testing/inference, we simplify the structure of CompenNeSt++ as shown in \autoref{fig:testing}. \textbf{(a)} Firstly, as mentioned in \autoref{subsec:warpingnet}, due to our novel cascaded coarse-to-fine network design and sampling strategy, WarpingNet can be substituted by a sampling grid and an image sampler shown as $ \T'^{-1} $ in \autoref{fig:testing}. \textbf{(b)} Secondly, CompenNeSt's surface feature extraction branch's (the top subnet of $ \F^{\dagger} $) weights and input are both fixed during testing, thus, it is trimmed and replaced by biases to reduce computation and memory usage. The biases are then directly applied to the  CompenNeSt backbone, we denote this simplified CompenNeSt++ as {\small$ \pi'^{\dagger}_{\vec{\theta}}$}. The two novel network simplification techniques make the proposed CompenNeSt++ both computationally and memory efficient with no performance drop.

	\subsection{Implementation and Training details}
	CompenNeSt++ is implemented using PyTorch \cite{paszke2017automatic} and trained using Adam optimizer \cite{kinga2015method} with a penalty factor of $ 10^{-4} $. The initial learning rate is set to $ 10^{-3} $ and decayed by a factor of 5 at the 1,000\textsuperscript{th} iteration. The model weights are initialized using the techniques in \autoref{subsec:domain_knowledge}. We train the model for 1,500 iterations on two Nvidia GeForce 1080Ti GPUs with a batch size of 48, and it takes about 15min to finish (without pre-trian).

	\subsection{Compensation pipeline}\label{subsec:pipeline}
	To summarize, our full projector compensation pipeline consists of three major steps.  \textbf{(1)} As shown in \autoref{fig:flowchart}(a), we start by projecting a plain gray image $ \x_0 $,  and $ N $ sampling images $ \x_1, \dots, \x_N $ to the projection surface and capture them using the camera, and denote the captured images as $ \tilde{\s} $ and $ \tilde{\x}_i $, respectively. \textbf{(2)} Then, we gather the $ N $ image pairs $\mathcal{X}=\{(\tilde{\x}_i, \x_i)\}_{i=1}^N $ and $ \tilde{\s} $ to train the compensation model {\small$ \pi^{\dagger}_{\vec{\theta}} = \{\F^{\dagger}_{\vec{\theta}}, \T^{-1}_{\vec{\theta}}\} $} end-to-end (see \autoref{fig:flowchart}(b)). \textbf{(3)} Afterwards, as shown in \autoref{fig:testing}, we simplify the trained CompenNeSt++ to {\small$ \pi'^{\dagger}_{\vec{\theta}}$} using the techniques in \autoref{subsec:cnn_testing}. Finally, for an ideal desired viewer-perceived image  $ \vec{z}$, we infer its compensation image $ \vec{z}^{*} $ and project $ \vec{z}^{*} $ to the surface.
	
	In practice, $ \vec{z} $ is physically restricted to the surface displayable area, \ie, a subregion of the projector FOV. Similar to \cite{raskar2003ilamps}, we find an optimal desired image $ \vec{z}' = \phi(\vec{z}; \vec{A}) $, where $ \vec{A} $ is a 2D affine transformation that uniformly scales and translates the ideal perceived image $ \vec{z} $ to optimally fit the projector FOV as shown in \autoref{fig:fov_mask} and \autoref{fig:testing}.
	
	%%%%%%%%%%%%%%%%%%%%%%%%%%%%%%%%%%%%%%%%%%%%%%%%%%%%%%%%%%%%%%%%%%%%%%%%%%%%
	\section{Benchmark}\label{sec:benchmark}
	An issue left unaddressed in previous studies is the lack of public benchmarks for quantitative evaluation, due mainly to the fact that traditional evaluation is highly setup-dependent. In theory, to evaluate a compensation algorithm, its output compensation image $\vec{x}^*$ for input $\vec{x}$ should be actually projected to the projection surface, and then captured by the camera and quantitatively compared with the ground truth. This process is obvious impractical since it requires the same projector-camera-environment setup for fair comparison of different algorithms.
	
	In this work, motivated by our problem formulation, we derive an effective surrogate evaluation protocol that requires no actual projection of the algorithm output. The basic idea is, according to \autoref{eq:training}, we can collect testing samples in the same way as the training samples. We can also evaluate an algorithm in the similar way.
	Specifically, we collect the test set of $M$ samples as $\mathcal{Y}=\{(\tilde{\vec{y}}_i, \vec{y}_i)\}_{i=1}^M$, under the same system setup as the training set $\mathcal{X}$. Then the algorithm performance is measured by averaging over similarities (\ie, PSNR, RMSE and SSIM \cite{wang2004image}) between each testing input image $\vec{y}_i$ and its algorithm output $\hat{\vec{y}}_i = \pi^{\dagger}_{\vec{\theta}}(\tilde{\vec{y}}_i; \tilde{\s})$ and reported in \autoref{tab:compare}.
	
	The above protocol allows us to construct a projector compensation evaluation benchmark, consisting of $K$ system setups, each with a training set $\mathcal{X}_k$, a test set $\mathcal{Y}_k$ and a surface image $\tilde{\vec{s}}_k$, $k=1,\dots,K$.
	
	\begin{figure}[!t]
		\begin{center}
			\includegraphics[width=1\linewidth]{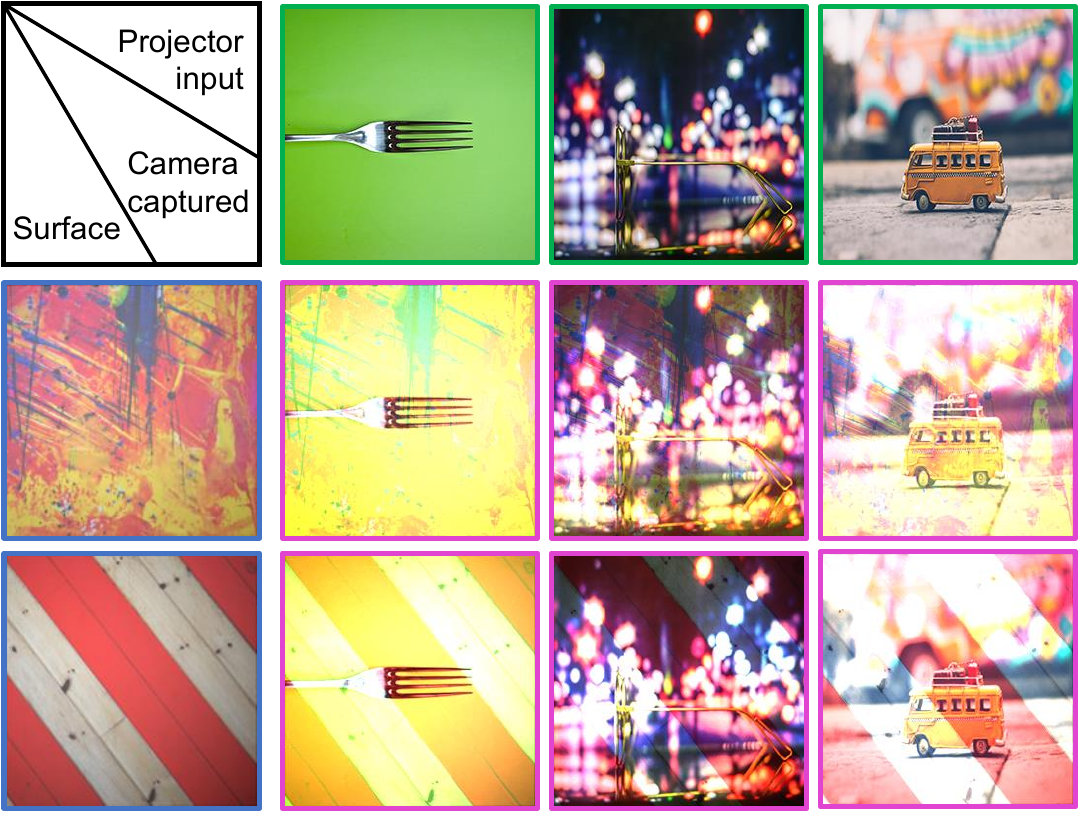}
			\caption{Blender rendered images (purple boxes). Two different surfaces (blue boxes) and three different projector input images (green boxes) are shown.
			}\label{fig:blender_sample}
		\end{center}
	\end{figure}

	\subsection{System configuration}
	Our projector compensation system consisted of a Canon 6D camera and a ViewSonic PJD7828HDL DLP projector with resolutions set to 640$\times$480 and 800$\times$600, respectively. In addition, an Elgato Cam Link 4K video capture card is connected to the camera to improve frame capturing efficiency (about 360ms per frame). Note that no serious radiometric calibration or camera color calibration are performed, instead for each setup, we manually adjust the camera by setting the picture style to faithful, and tuning  	exposure and white balance such that the camera-captured image roughly matches human perception.
	
	The distance between the camera and the projector varies in the range of 500mm to 1,000mm and the projection surface is around 1,000mm in front of the projector-camera pair. The camera exposure, focus and white balance modes are set to manual; and the global lighting varies for each setup but is fixed during each setup's data capturing and system testing.
	
	\subsection{Dataset}
	\subsubsection{Real data}
	To obtain the sampling colors and textures as diverse as possible, we downloaded 700 colorful textured images from the Internet and use $ N=500 $ for each training set $\mathcal{X}_k$ and $ M = 200$ for each testing set $\mathcal{Y}_k$. 
	
	\noindent\textbf{Full compensation dataset.}
	In total $K = 20$ different setups are prepared for training and evaluation, each setup has a \emph{nonplanar} textured surface. 
	Future works can replicate our results and compare with CompenNeSt++ on the benchmark without replicating our setups. 

	\noindent\textbf{Photometric compensation dataset.}
	In addition, considering that some projector-camera systems focus on planar surfaces, we propose a photometric compensation dataset with total $24$ different setups for photometric compensation evaluations. This dataset consists of \emph{planar} textured surfaces only and the geometry is corrected using a homography. This dataset is particularly useful for analyzing the photometric process by removing the geometric process involved in WarpingNet. In fact, to abbreviate potential geometry disturbances we investigate CompenNeSt architecture in an ablation study on this dataset, as shown in \autoref{tab:compennest_explore}, \autoref{fig:abl} and \autoref{fig:act}.

	\subsubsection{Synthetic data}\label{subsec:synthetic_data}
	As mentioned in \autoref{subsec:pre-train}, we propose a pre-training method to improve practicability of CompenNeSt++. Intuitively, it is better to pre-train the model on real data, however, it is difficult and time consuming to capture a real dataset consisting of a wide range of setup parameter variations, such as different lighting, surface material, projector-camera settings, and poses, etc. 
	Instead, we synthesized a dataset using Blender and the virtual projector-camera system consisted of \textit{a camera}, \textit{a projector} and \textit{a textured surface}. As our WarpingNet (the geometric part) is setup-dependent and cannot be pre-trained, we restrict the surface to be planar and focus on photometric compensation \ie, CompenNeSt only. The surface material is modeled using a tunable principled BSDF shader. For each setup, the surface base color is set with different textured images. To increase diversity, random perturbations are applied to the camera parameters and the camera/projector/surface poses. 
	
	In total, we render 100 synthetic setups and each setup consists of 500 image pairs for training and 200 for testing. Some representative samples are shown in \autoref{fig:blender_sample}. Note that in this dataset, both the surface patterns and the projected sampling images are different from the real dataset. Future works can leverage this dataset for model pre-training and network architecture exploration.

	\begin{table*}[!t]
		\caption{Quantitative comparison of full compensation algorithms. Results are averaged over $K= 20$ different setups.  
		Note that the metrics for uncompensated images are PSNR=9.5973, RMSE=0.5765 and SSIM=0.0767. The metrics for the original TPS \cite{grundhofer2015robust} w/ SL (\#Train=125) are PSNR=16.7271, RMSE= 0.2549 and SSIM=0.5207. See \href{https://vision.cs.stonybrook.edu/~bingyao/pub/CompenNeSt_supp}{supplementary material} for separate measurements for each setup.}\label{tab:compare}
		\begin{center}
			{\footnotesize 
				\begin{tabular}{@{}l@{\hspace{5mm}}c@{\hspace{2mm}}c@{\hspace{2mm}}c@{\hspace{5mm}}c@{\hspace{2mm}}c@{\hspace{2mm}}c@{\hspace{5mm}}c@{\hspace{2mm}}c@{\hspace{2mm}}c@{\hspace{5mm}}c@{\hspace{2mm}}c@{\hspace{2mm}}c@{}}
					\toprule
					\rowcolor[gray]{1}
					\multicolumn{1}{c}{} & \multicolumn{3}{c}{\textbf{\#Train=48}}                & \multicolumn{3}{c}{\textbf{\#Train=125}}               & \multicolumn{3}{c}{\textbf{\#Train=250}}               & \multicolumn{3}{c}{\textbf{\#Train=500}}               \\ 
					\multicolumn{1}{c}{\multirow{-2}{*}{\cellcolor[gray]{1}\textbf{Model}}}                                &  \textbf{PSNR}$\uparrow$ & \textbf{RMSE}$\downarrow$ & \textbf{SSIM}$\uparrow$ & \textbf{PSNR}$\uparrow$ & \textbf{RMSE}$\downarrow$ & \textbf{SSIM}$\uparrow$ & \textbf{PSNR}$\uparrow$ & \textbf{RMSE}$\downarrow$ & \textbf{SSIM}$\uparrow$ & \textbf{PSNR}$\uparrow$ & \textbf{RMSE}$\downarrow$ & \textbf{SSIM}$\uparrow$ \\\midrule
					TPS \cite{grundhofer2015robust} textured w/ SL                                  & 18.0297       & 0.2199        & 0.5390        & 18.0132       & 0.2205        & 0.5687        & 18.0080       & 0.2206        & 0.5787        & 17.9746       & 0.2215        & 0.5830        \\
					Pix2pix \cite{isola2017image} w/ SL                                       & 17.7160       & 0.2271        & 0.5068        & 17.1141       & 0.2468        & 0.5592        & 16.5236       & 0.2669        & 0.5763     &   19.4544 & 0.1893 & 0.6222       \\		
					CompenNet++ \cite{huang2019compennet++}                                     & {19.8552}       & {0.1781}        & {0.6637}        & {20.7947}       & {0.1598}        & {0.7116}        & {20.8959}       & {0.1581}        & {0.7227}        & {21.1127}       & {0.1540}        & {0.7269}        \\
					CompenNeSt w/SL            & \textbf{20.2788}	 &\textbf{0.1708}       & \textbf{0.6890}         &  \textbf{21.0508}          & \textbf{0.1560}	       & \textbf{0.7219}   & \textbf{21.3389}          & \textbf{0.1508} & 0.7376   &\textbf{21.5184}	         & \textbf{0.1476}       & 0.7413        \\	
					CompenNeSt++ w/o surf.           &  18.1238	&0.2168	&0.6195 &18.9314	&0.1974	&0.6623  &19.1256	&0.1930	&0.6739 & 19.2202	&0.1909	&0.6754\\ 
					CompenNeSt++ w/o refine           & {19.3868} & {0.1934} & {0.6372} 	& {20.7373} & {0.1614} & {0.7092} & {21.0232} & {0.1561} &{0.7246} 		
					& {21.2691} & {0.1516} & {0.7322} \\ 	
					CompenNeSt++           & {19.9077} & {0.1775} & {0.6764} & {20.8597} & {0.1590} & {0.7202} & {21.2496} & {0.1518} & \textbf{0.7393} & {21.4868} & {0.1477} & \textbf{0.7468} \\ \bottomrule
				\end{tabular}
			}		
		\end{center}
	\end{table*}
	
		\begin{table*}[!t]
		\caption{Quantitative comparison between  CompenNeSt w/SL and CompenNeSt++. Results are averaged over $K= 2$ setups with \emph{specular highlight} surfaces. CompenNeSt++ clearly performs better in this particular case. }\label{tab:compare_spec}
		\begin{center}
			{\footnotesize 
				\begin{tabular}{@{}l@{\hspace{5mm}}c@{\hspace{2mm}}c@{\hspace{2mm}}c@{\hspace{5mm}}c@{\hspace{2mm}}c@{\hspace{2mm}}c@{\hspace{5mm}}c@{\hspace{2mm}}c@{\hspace{2mm}}c@{\hspace{5mm}}c@{\hspace{2mm}}c@{\hspace{2mm}}c@{}}
					\toprule
					\rowcolor[gray]{1}
					\multicolumn{1}{c}{} & \multicolumn{3}{c}{\textbf{\#Train=48}}                & \multicolumn{3}{c}{\textbf{\#Train=125}}               & \multicolumn{3}{c}{\textbf{\#Train=250}}               & \multicolumn{3}{c}{\textbf{\#Train=500}}               \\ 
					\multicolumn{1}{c}{\multirow{-2}{*}{\cellcolor[gray]{1}\textbf{Model}}}                                &  \textbf{PSNR}$\uparrow$ & \textbf{RMSE}$\downarrow$ & \textbf{SSIM}$\uparrow$ & \textbf{PSNR}$\uparrow$ & \textbf{RMSE}$\downarrow$ & \textbf{SSIM}$\uparrow$ & \textbf{PSNR}$\uparrow$ & \textbf{RMSE}$\downarrow$ & \textbf{SSIM}$\uparrow$ & \textbf{PSNR}$\uparrow$ & \textbf{RMSE}$\downarrow$ & \textbf{SSIM}$\uparrow$ \\\midrule
					CompenNeSt w/SL            & 17.1685 &	0.2433 &	0.5021
         &  18.1294 &	0.2173	 & 0.5569
   & 18.6915 &	0.2034 &	0.5881
   &18.9969 &	0.1960 &	0.5966
        \\	
					CompenNeSt++           & \textbf{17.5909} &	\textbf{0.2302} &	\textbf{0.5444}
 & \textbf{18.5610} &	\textbf{0.2054} &	\textbf{0.6092}
 & \textbf{19.1894}  &	\textbf{0.1916} &	\textbf{0.6415}
 & \textbf{19.4663} &	\textbf{0.1851} &	\textbf{0.6459}
 \\ \bottomrule
				\end{tabular}
			}		
		\end{center}
	\end{table*}

	%%%%%%%%%%%%%%%%%%%%%%%%%%%%%%%%%%%%%%%%%%%%%%%%%%%%%%%%%%%%%%%%%%%%%%%%%%%%
	\section{Experimental Evaluations}\label{sec:experiments}
	
	\subsection{Comparison with state-of-the-arts}\label{subsec:comparison_existing}
	We compare the proposed full compensation method (\ie, CompenNeSt++) with four two-step baselines, a context-independent TPS\footnote{Not geometric correction \cite{donato2002approximate}, instead using TPS to model the pixel-wise \emph{photometric compensation function}.} model \cite{grundhofer2015robust}, an improved TPS model (explained below), a Pix2pix \cite{isola2017image} model and a CompenNeSt model that is without WarpingNet on the proposed evaluation benchmark.
				
	To fairly compare two-step methods, we use the same structured light (SL) warping for geometric correction. We first projected 42 Gray-coded SL patterns \cite{moreno2012simple} to establish projector-camera pixel-to-pixel mapping. Due to strong photometric disturbance, the SL method \cite{moreno2012simple} might suffer from decoding errors and thus we use bilinear interpolation to fill the missing correspondences. Afterwards, we capture 125 pairs of plain color sampling image as used in the original TPS method~\cite{grundhofer2015robust} for photometric compensation, then we warp the sampling images to the projector canonical frontal view using SL and name this method \emph{TPS w/ SL}. We also fit the TPS method using SL-warped diverse \emph{textured} training set $\mathcal{X}_k$, and name this method \emph{TPS textured w/ SL}.
	
	The experiment results in \autoref{tab:compare} show clear improvement of TPS textured over the original TPS method. Our explanations are: \textbf{(a)} Compared with plain color images, the textured training images and testing images share a more similar distribution. \textbf{(b)} Although the original TPS method uses $ 5^3 $ plain color images, each projector pixel's R/G/B channel only has five different intensity levels, training the TPS model using these samples may lead to a suboptimal solution. While our colorful textured samples evenly cover the RGB space at each projector pixel, resulting a more faithful sampling of the photometric compensation function.
	
	\begin{figure*}[!t]
		\begin{center}
			\includegraphics[width=1\linewidth]{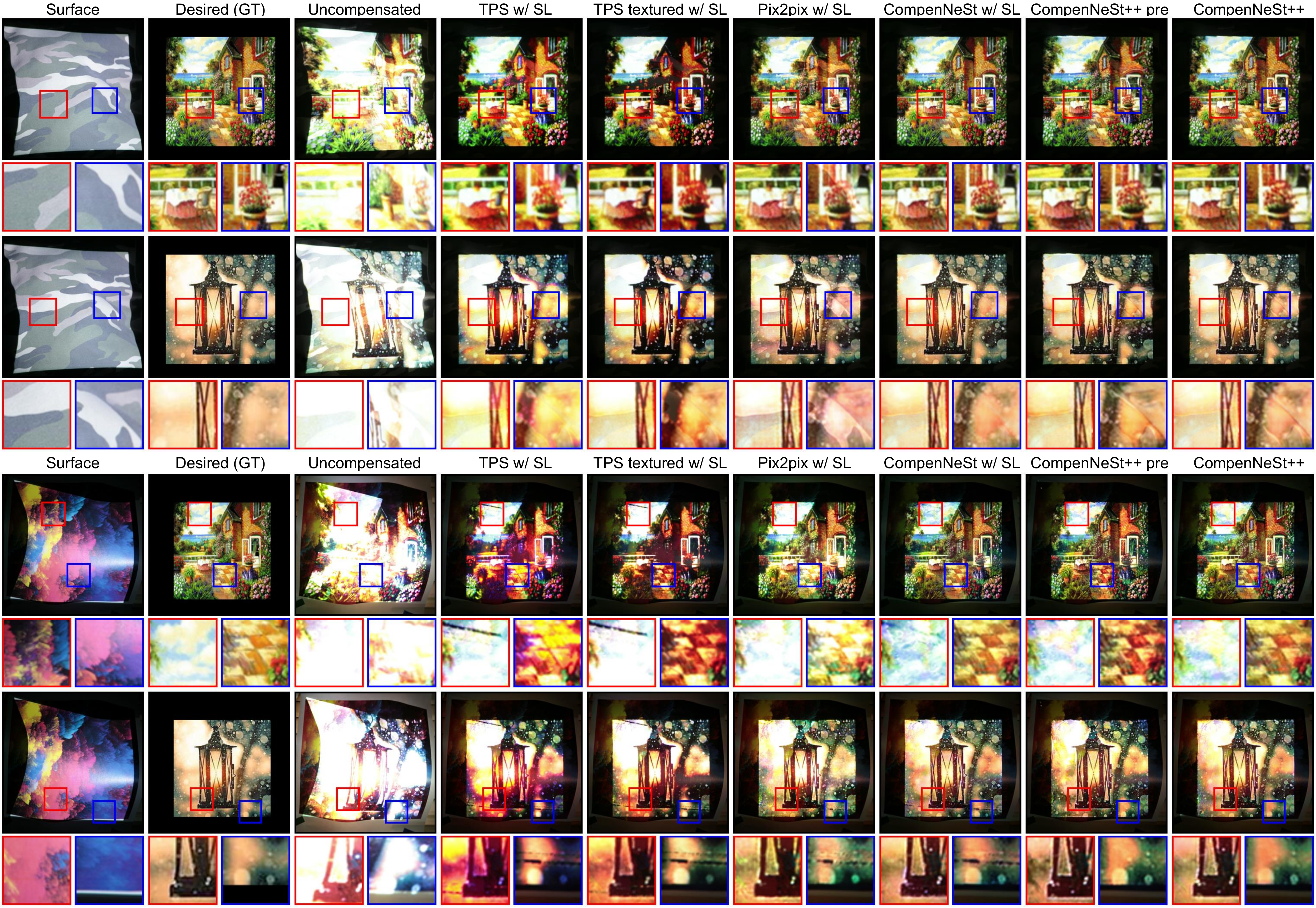}
			\caption{Qualitative comparison of TPS \cite{ grundhofer2015robust} w/ SL, TPS textured w/ SL, Pix2pix \cite{isola2017image} w/ SL, our CompenNeSt w/ SL, our pre-trained CompenNeSt++ fine-tuned using only 8 sampling images, \ie, \emph{CompenNeSt++ pre} and our CompenNeSt++ on two different surfaces. All models were trained using 500 sampling images (except for \emph{CompenNeSt++ pre}). 
			The 1\textsuperscript{st} to 3\textsuperscript{rd} columns are camera-captured projection surface, desired viewer-perceived image and camera-captured uncompensated projection, respectively. The rest columns are compensation results of different methods. Each image is provided with two zoomed-in patches for detailed comparison. See \href{https://vision.cs.stonybrook.edu/~bingyao/pub/CompenNeSt_supp}{supplementary material} for more results.
				\label{fig:compare_real}
			}
		\end{center}
	\end{figure*}

	To demonstrate the difficulty of full compensation problem, we compare with a deep learning-based image-to-image translation model Pix2pix\footnote{{\scriptsize \url{https://github.com/junyanz/pytorch-CycleGAN-and-Pix2pix}}} \cite{isola2017image} trained on the same SL-warped $\mathcal{X}_k$ as TPS textured w/ SL, we named it \emph{Pix2pix w/ SL}. We train Pix2pix for 12,000 iterations to match the training time of our model. The results show that the proposed CompenNeSt++ outperforms Pix2pix w/ SL, demonstrating that the full compensation problem cannot be well solved by a general deep-learning based image-to-image translation model.
	
	We then compare our method with our partial compensation model CompenNeSt and we train it with the same SL-warped training set $\mathcal{X}_k$ as TPS textured w/ SL and Pix2pix w/ SL, and name this two-step method \emph{CompenNeSt w/ SL}. The quantitative and qualitative comparisons are shown in \autoref{tab:compare} and \autoref{fig:compare_real}, respectively.

	\autoref{tab:compare} clearly shows that CompenNeSt++ outperforms other two-step methods, except for CompenNeSt w/ SL.
	This indicates that even without an additional structured light step, the geometry correction can be learned directly from the photometric sampling images. 
	
	Note that SL may not work well on surfaces with specular highlight, \eg, as shown in \autoref{tab:compare_spec} that CompenNeSt++ outperforms CompenNeSt w/ SL by a significant margin on two specular highlight setups extracted from the 20 full compensation setups in \autoref{tab:compare}. 
	This is because SL suffers from decoding error due to specular highlight and 
	solving full compensation problem separately may lead to suboptimal solution, and thus the two steps should be solved jointly, as proposed by CompenNeSt++. Besides outperforming CompenNeSt w/ SL on specular highlight surfaces, CompenNeSt++ uses 42 fewer images than two-step SL-based methods. 
	
	We explain why two-step methods may find suboptimal solution in \autoref{fig:compare_real}, where SL decoding errors affect the photometric compensation accuracy. In the 3\textsuperscript{rd} row red zoomed-in patches and the 4\textsuperscript{th} row blue zoomed-in patches, we see unfaithful compensations by the SL-based two-step methods (4\textsuperscript{th}-7\textsuperscript{th} columns), because SL suffers from decoding errors due to specular reflections and establishes false pixel mappings. Then, a second step of photometric compensation based on a false mapping is inevitably error prone.
	On the contrary, this issue is better addressed by the proposed end-to-end methods \emph{CompenNeSt++ pre} (\ie, pre-trained and fine-tuned using only 8 sampling images) and \emph{CompenNeSt++} (last two columns), where global geometry and photometry information is considered in full compensation and gradients of the image reconstruction loss can be backpropagated to both modules. In summary, CompenNeSt++ not only brings improved performance than two-step SL-based methods, but also waives 42 extra SL projections/captures, and meanwhile being insensitive to specular highlights. Moreover, the pre-trained model \emph{CompenNeSt++ pre} can work with only 8 sampling images, which further adds to the advantages of our method.

	%%%%%%%%%%%%%%%%%%%%%%%%%%%%%%%%%%%%%%%%%%%%%%%%%%%%%%%%%%%%%%%%%%%%%%%%%%%%
	\subsection{Ablation study}	
	In this section, we conduct various ablation studies to show the effectiveness of our novel end-to-end problem formulation, network architecture and further analyze the mechanism of deep projector compensation.

	\subsubsection{Network architecture exploration}\label{subsec:structure_explore}
	Below we show how we explore the proposed CompenNeSt (the photometric part of CompenNeSt++) architecture and compare it with its degraded versions and our previous photometric compensation model CompenNet \cite{huang2019compennet} on the \emph{photometric compensation benchmark} to show the effectiveness of our network design. Then, we compare CompenNeSt++ with our previous full compensation model CompenNet++ \cite{huang2019compennet++} on the \emph{full compensation benchmark} to show that by incorporating the improved CompenNeSt, CompenNeSt++ significantly outperforms CompenNet++ \cite{huang2019compennet++} (see \autoref{tab:compare}).

	\begin{figure*}[!t]
		\begin{center}
			\includegraphics[width=1\linewidth]{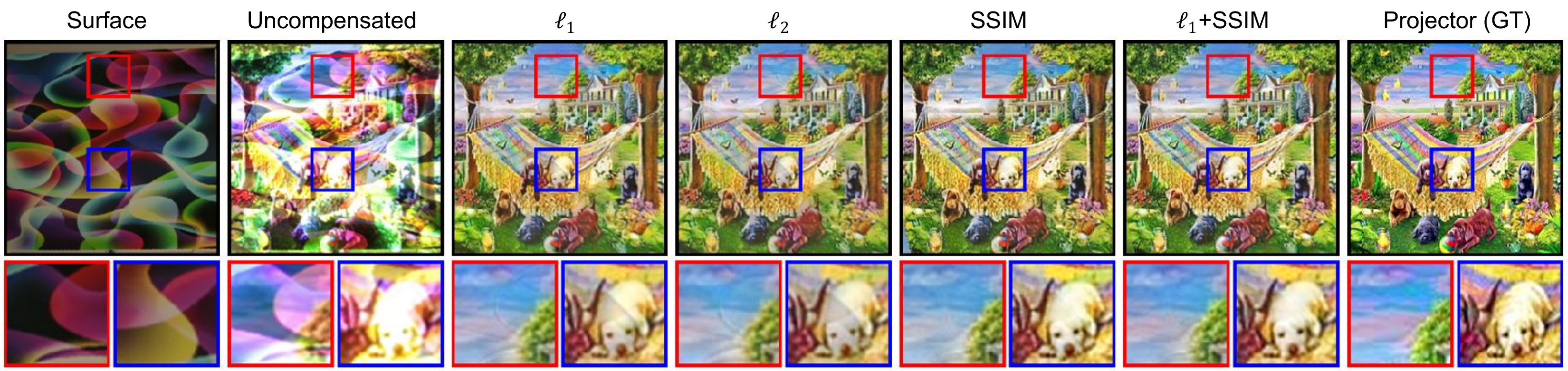}
			\caption{Qualitative comparison of CompenNeSt++ trained with $ \ell_1$ loss, $\ell_2$ loss, SSIM loss and $ \ell_1 +$SSIM loss. Clearly, $ \ell_1 $ and $ \ell_2 $ losses are unable to successfully compensate the surface patterns (see the dog head). $ \ell_1 +$SSIM and the SSIM losses produce similar results, but the cloud in the red zoomed-in patch of SSIM is grayer than $ \ell_1 +$ SSIM and the ground truth.}\label{fig:compare_loss}
		\end{center}
	\end{figure*}

	\noindent\textbf{Effectiveness of the siamese structure and improved layers.} 
	Compared with CompenNet \cite{huang2019compennet}, CompenNeSt has three major improvements: (1) a novel siamese structure (for the orange encoder part, CompenNet does not share weights); (2) symmetric skip connections and thus surface feature \emph{subtraction} can be performed; (3) replacing the 2$\times$2 filter of the first transposed convolutional layer to a 3$\times$3 filter. 
	To show the effectiveness of our new architecture, we compare with three respective degraded versions, \ie, the original \emph{CompenNet} \cite{huang2019compennet}, \emph{CompenNeSt (2$\times$2)} and \emph{CompenNeSt (2$\times$2, deg. skip)} (deg. skip means using degraded CompenNet-like skip connections, the only difference between this model and CompenNet \cite{huang2019compennet} is its siamese structure).

	As the quantitative comparisons shown in \autoref{tab:compennest_explore}, the proposed CompenNeSt outperforms all other degraded versions, demonstrating the effectiveness of the siamese structure, the feature subtraction operation and improved layers.

	\noindent\textbf{Effectiveness of the surface image.}
	To show the effectiveness of our learning-based formulation in \autoref{eq:training} and that the surface image $ \tilde{\vec{s}} $ is a necessary model input, we compare the proposed CompenNeSt/CompenNeSt++ with their degraded versions that are without the input surface image and the corresponding encoder branch. We named them \emph{CompenNeSt w/o surf.} in \autoref{tab:compennest_explore} and \emph{CompenNeSt++ w/o surf.} in \autoref{tab:compare}. Clearly, CompenNeSt and CompenNeSt++ outperform their degraded versions that are without the surface input on the photometric compensation and full compensation benchmark, respectively.

	In particular, in \autoref{tab:compennest_explore} we can see clear improved PSNR/RMSE/SSIM  when $ \tilde{\vec{s}} $ is included in the model input, showing that our learning-based formulation has a clear advantage over the models that ignore the important information encoded in the surface image. Secondly, in \autoref{tab:compare} \emph{CompenNeSt++ w/o surf.} outperforms TPS w/ SL and TPS textured w/ SL and Pix2pix w/ SL on PSNR/RMSE/SSIM even when $ \tilde{\vec{s}} $ is not included, showing the effectiveness of context-dependent formulation and the importance of the task-specific network design and the problem domain knowledge.

	\noindent\textbf{Effectiveness of the grid refinement network.}
	To demonstrate the effectiveness of the sampling grid refinement network $\mathcal{W}_{\vec{\theta}\textsubscript{r}}$ (\autoref{eq:grid_sampling} and \autoref{fig:flowchart}), we created a degraded CompenNeSt++ by removing $\mathcal{W}_{\vec{\theta}\textsubscript{r}}$, and name it \emph{CompenNeSt++ w/o refine}. As reported in \autoref{tab:compare}, CompenNeSt++ clearly outperforms this degraded model, showing the effectiveness of the grid refinement network $\mathcal{W}_{\vec{\theta}\textsubscript{r}}$.

	\begin{table}[!t]
		\begin{center}
			\caption{Quantitative comparison of the proposed \emph{CompenNeSt} with CompenNet \cite{huang2019compennet} and three degraded versions that are (1) without the surface image, (2) with CompenNet-like 2$\times$2 transposed convolutional filters; and (3) additionally with CompenNet-like degraded skip convolutional layers. The models are compared on the \textbf{photometric compensation dataset} using 500 images and 1,000 iterations and the results are averaged over $ K=24 $ setups.}\label{tab:compennest_explore}
			{\small \begin{tabular}[]{lccc}
					\toprule
					\rowcolor[gray]{1}
					\textbf{Model} & \textbf{PSNR}$\uparrow$ & \textbf{RMSE}$\downarrow$ & \textbf{SSIM}$\uparrow$\tabularnewline
					\midrule
					CompenNet \cite{huang2019compennet} & 21.7998 & 0.1425 & 0.7523\\
					CompenNeSt w/o surf. & 20.6123 & 0.1633 &	0.7319\tabularnewline
					CompenNeSt (2$\times$2)  & 22.1100 & 0.1373 &	0.7698\tabularnewline
					CompenNeSt (2$\times$2, deg. skip)  & 21.9101	&0.1404&	0.7595\tabularnewline		
					CompenNeSt & \textbf{22.2992}	& \textbf{0.1347}& \textbf{0.7753}\tabularnewline \midrule
					Uncompensated & 12.1673 & 0.4342 & 0.4875\tabularnewline
					\bottomrule
			\end{tabular}}
		\end{center}
	\end{table}

	\subsubsection{Comparison of different loss functions}\label{subsec:loss}
	Previous methods fit the composite compensation function by a pixel-wise $ \ell_2 $ loss and it is known to penalize large pixel errors while ignores the structural details \cite{wang2004image,zhao2017loss}. We investigated four different loss functions, i.e., pixel-wise $ \ell_1 $ loss, pixel-wise $ \ell_2 $ loss, SSIM loss, and  $ \ell_1 +$SSIM loss. The qualitative and quantitative comparisons are shown in \autoref{fig:compare_loss} and \autoref{tab:compare_loss}, respectively. In \autoref{fig:compare_loss}, compared with SSIM and $ \ell_1 +$SSIM losses, pixel-wise $ \ell_1 $ and $ \ell_2 $ losses cannot well compensate surface patterns, as shown by the dog head in the blue zoomed-in patches. Compared with $ \ell_1 +$SSIM loss, SSIM loss cannot well compensate the color as shown by the cloud in the red zoomed-in patches. 
	
	The quantitative comparisons in \autoref{tab:compare_loss} are also consistent with the qualitative comparisons in in \autoref{fig:compare_loss}. Note that SSIM loss alone obtains a worse PSNR/RMSE than $ \ell_1 $ and $ \ell_2 $ losses and a worse SSIM value than  $ \ell_1 +$SSIM because it failed to converge on some setups with hard surface geometries and the output becomes plain gray. We further investigated the issue and found that compared with pixel-wise $ \ell_1 $ and $ \ell_2 $ losses, SSIM loss alone might encourage smooth plain gray patches. This problem also exists when we train with very few sampling images (see \autoref{subsec:exp_pre-train}). Thus, we use $ \ell_1 +$SSIM loss for CompenNeSt++ training. 
	
	Moreover, even when trained with pixel-wise $ \ell_1 $ loss, CompenNeSt++ outperforms TPS, TPS textured and Pix2pix on PSNR, RMSE and SSIM, this again shows a clear advantage of our task-targeting formulation and architecture.

\begin{table}[!t]
	\begin{center}
	\caption{Quantitative comparison of different loss functions for the proposed \emph{CompenNeSt++} on the \textbf{full compensation dataset} using 500 images and 1,500 iterations and the results are averaged over $ K=20 $ setups.  }\label{tab:compare_loss}
		{\small \begin{tabular}[]{lccc}
				\toprule
				\rowcolor[gray]{1}
				\textbf{Loss} & \textbf{PSNR}$\uparrow$ & \textbf{RMSE}$\downarrow$ & \textbf{SSIM}$\uparrow$\tabularnewline
				\midrule					
				$ \ell_1 $ &20.9856&	0.1564&	0.6819\tabularnewline
				$ \ell_2 $ &20.4036&	0.1669&	0.6523\tabularnewline
				SSIM & 20.1489 &	0.2001 &	0.7213\tabularnewline
				$ \ell_1 +$SSIM &\textbf{21.4868}&	\textbf{0.1477}&	\textbf{0.7468} \tabularnewline\midrule
				Uncompensated &9.5973& 0.5765 & 0.0767\tabularnewline
				\bottomrule
		\end{tabular}}
	\end{center}
\end{table}

\begin{figure*}[!t]
	\centering    
	\includegraphics[width=1\linewidth]{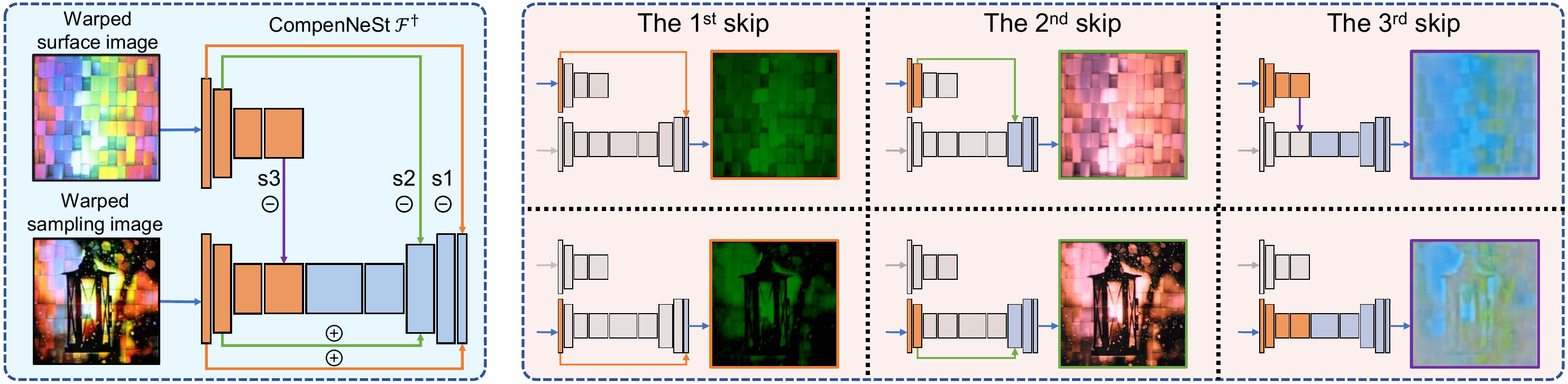}
	  \caption{Visualization of CompenNeSt photometric compensation mechanism. \textbf{Left}: a \emph{trained} CompenNeSt takes two warped images as input and we investigate the feature maps by enabling the input and the corresponding skip connections once at a time. \textbf{Right}: the top and the bottom rows show the network outputs when input the surface image or the sampling image, respectively; and each column shows the output when a specific skip connection and the corresponding layers are enabled. We use gray color to indicate \emph{disabled} inputs, modules and connections. As shown in the first two columns, the feature maps of the first two layers carry low-level texture information and green/red components. In the 3rd column, we see that the feature maps of the fourth layer carry high-level global information and blue and yellow components.}
	  \label{fig:act} 
	\end{figure*}

	\subsubsection{Interpretation of CompenNeSt photometric compensation mechanism} \label{subsec:ablation_act}
	To interpret the  photometric compensation mechanism of CompenNeSt, we conduct two ablation studies. 
	
	First, we investigated the features carried by each of the three skip connections by enabling the surface/sampling image features and their corresponding skip connections \emph{one-by-one} and plot the model output in \autoref{fig:act}. 
	
\begin{figure*}[!t]
			\centering    
			\includegraphics[width=1\linewidth]{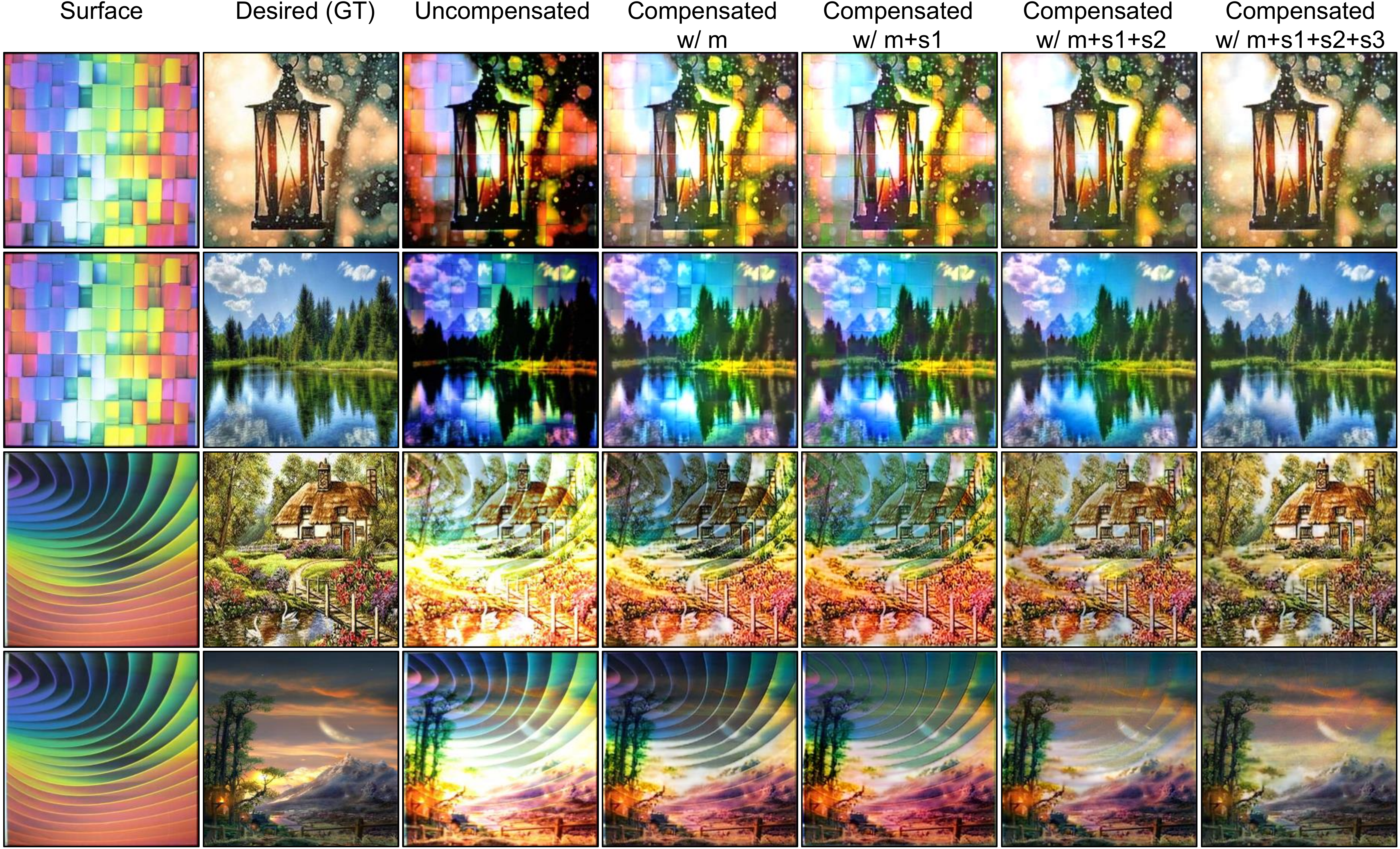}          
			\caption{Output of CompenNeSt when \textbf{sequentially} enabling the three \emph{surface} skip connections. We start with a \emph{trained} CompenNeSt and an uncompensated camera-captured sampling image and disable all the skip connections between the surface branch and the backbone network (\ie, s1-s3 in \autoref{fig:act}). Then, we sequentially enabled s1 to s3 as shown in columns 4-7. Note that after we disabled a surface skip connection, we subtract its feature \textbf{mean}, \eg, \emph{Compensated w/ m} means that we disabled s1-s3 but subtracted their corresponding surface feature means from the backbone network. Compared with subtracting the actual feature map, subtracting feature mean only performs a global color and brightness adjustment (see the difference between the 3rd and the 4th columns). Then, when we enabled a surface skip connection, the feature \textbf{variance}/texture information can be better visualized. \Eg, comparing the 4th with the 5th/6th columns, we see that s1 and s2 carry low-level surface texture features, subtracting them significantly removes the surface pattern. Comparing the 6th and the 7th columns, we see that s3 carries global color information.}
			\label{fig:abl} 
		\end{figure*}

	Then, we show how the surface pattern was gradually compensated by \emph{sequentially subtracting} the surface features from the sampling image features via the three skip connections. The network outputs are shown in \autoref{fig:abl} columns 4-7. Note that unlike \autoref{fig:act} where the three skip connections are enabled one at a time, in \autoref{fig:abl}, the three skip connections are gradually enabled, showing how the output (the 7th column) is gradually compensated by subtracting the three surface features. 	
	
		\begin{table}[!t]
			\begin{center}
				\caption{Quantitative comparison between the pre-trained \emph{CompenNeSt++} and the default \emph{CompenNeSt++} on the full compensation benchmark. Both models were trained using \emph{only 8 samples and 800 iterations with a batch size of 8, and took about 3 minutes}. }\label{tab:pretain}
				{\small \begin{tabular}[]{lccc}
						\toprule
						\rowcolor[gray]{1}
						\textbf{Model} & \textbf{PSNR}$\uparrow$ & \textbf{RMSE}$\downarrow$ & \textbf{SSIM}$\uparrow$\tabularnewline
						\midrule
						CompenNeSt++ & 16.6975 &	0.2584	& 0.5112 \tabularnewline
						CompenNeSt++ pre-trained  & \textbf{18.4551} &	\textbf{0.2108} &	\textbf{0.6410} \tabularnewline
						\midrule	
						Uncompensated &9.5973& 0.5765 & 0.0767\tabularnewline
						\bottomrule
				\end{tabular}}
			\end{center}
		\end{table}

	\subsubsection{Practicability of the pre-training method}\label{subsec:exp_pre-train}
	As mentioned before (\autoref{subsec:pre-train}), full projector compensation need to be quickly retrained when setup changes, however all existing methods must \emph{rerun the projection-capturing-compensation process from scratch}, limiting practicability of projector-camera systems. Below we show our pre-trained CompenNeSt++ can achieve good quality even when only 8 sampling images are available.

	We compare our default CompenNeSt++ trained from scratch with a CompenNeSt++ pre-trained on our Blender rendered synthetic dataset. Then we train/fine-tune and evaluate both models on the full compensation benchmark. To demonstrate that the pre-trained model improves performance with limited training pairs and training time, we trained both models for 800 iterations using only 8 training pairs. Note that for limited training images, $ \ell_1 $+SSIM loss may lead to instability to the WarpingNet training, thus we use $ \ell_1 $ loss for the first 200 iterations and switch to $ \ell_1 $+SSIM loss for the remaining 600 iterations. The comparison results are reported in \autoref{tab:pretain} and \autoref{fig:compare_real}.
	
	Clearly, we see that the pre-trained CompenNeSt++ outperforms the default CompenNeSt++, even though the 20 real training and evaluation setups have different projector-camera settings, sampling images and surface patterns as the synthetic pre-trained setup. Our explanation is that despite the different settings, the pre-trained model can still learn the \emph{compensation operation}, \ie, surface feature subtraction/pattern cancelling  (\autoref{subsec:compennet}) from 100 different synthetic setups.  This pre-trained model makes our method particularly practical for new setups, because it can be quickly tuned with much fewer training images and thus shortens the image capturing and training time. 
	
	Moreover, even with limited 8 training pairs and 800 iterations, the pre-trained CompenNeSt++ outperforms TPS \cite{grundhofer2015robust}, TPS textured and Pix2pix \cite{isola2017image} trained with \emph{250 images} on PSNR/RMSE/SSIM in \autoref{tab:compare}.
	
	Furthermore, CompenNeSt++ has much fewer parameters (0.8M) than Pix2pix's default generator (54M parameters). This further confirms that projector compensation is a complex problem and is different from general image-to-image translation tasks, and carefully designed models and domain knowledge are necessary.

	%%%%%%%%%%%%%%%%%%%%%%%%%%%%%%%%%%%%%%%%%%%%%%%%%%%%%%%%%%%%%%%%%%%%%%%%%%%%
	\section{Conclusions and Limitations}\label{sec:conclusions}
	In this paper, for the first time, we reformulate the full projector compensation problem as a learning problem and propose an accurate and practical end-to-end solution named CompenNeSt++. In particular, CompenNeSt++ jointly learns geometric correction and photometric compensation without an additional structure light step, thus being end-to-end differentiable and waiving 42 extra SL images.
	The effectiveness of our formulation and architecture is verified by comprehensive experimental evaluations and ablation studies. Moreover, for the first time, we provide the community with two novel setup-independent evaluation benchmark datasets. Our method is evaluated carefully on the benchmarks, and the results show that our end-to-end learning solution outperforms state-of-the-arts both qualitatively and quantitatively by a significant margin. To make our model more practical, we propose a synthetic dataset and a pre-training method, which allows our model to adapt to new setups with only 8 images and shorter training time, adding to the advantages over the prior works.

\begin{figure}[!t]
		\begin{center}
			\includegraphics[width=1\linewidth]{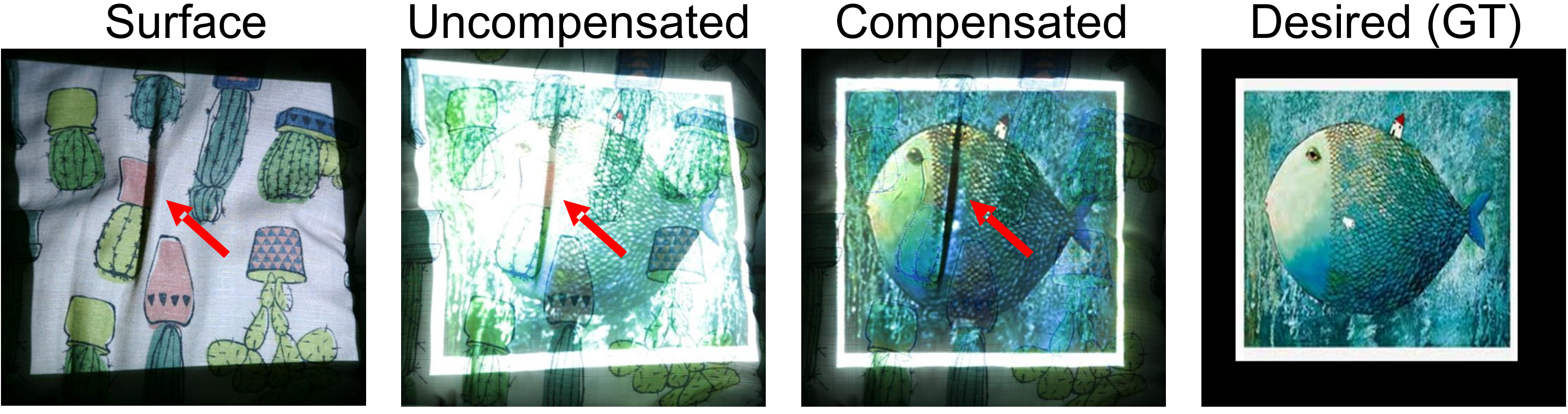}
			\caption{CompenNeSt++ is unable to compensate occluded regions such as the pillow fold as pointed by the red arrows.}\label{fig:failed}
		\end{center}
	\end{figure}

	\noindent\textbf{Limitations and future work.} Our WarpingNet may not work for surfaces with hard edges. We assume that each single patch of the projection surface can be illuminated by the projector. That said, CompenNeSt++ may not work well on complex surfaces with occlusions (\autoref{fig:failed}). One potential solution is to use multiple projectors to cover each other's blind spots. In fact, extending the end-to-end full compensation framework to multiple projectors is an interesting future direction. 
	
	For each setup, instead of serious camera color calibration, we manually adjust the camera exposure and white balance, and set the picture style to faithful mode to roughly match human perception. The  camera-captured images also contain projector and the camera lens distortions, which are jointly learned with the projection surface distortions by WarpingNet during network training. Thus, the camera-captured results may not perfectly reflect real human perceived effects. Moreover, CompenNeSt++ only works for \textbf{static} setups, and requires retraining when setup changes (\eg, moving/replacing surfaces, moving the projector or the camera/viewer's angle). The proposed benchmark is also built for static full projector compensation. Extending the proposed method to dynamic projection-mapping is definitely a promising direction in future work.
	
% \appendices
% \section{test}

% % use section* for acknowledgment
% \ifCLASSOPTIONcompsoc
%   % The Computer Society usually uses the plural form
%   \section*{Acknowledgments}
% \else
%   % regular IEEE prefers the singular form
% \fi

\ifCLASSOPTIONcaptionsoff
  \newpage
\fi

%%%%%%%%%%%%%%%%%%%%%%%%%%%%% bib %%%%%%%%%%%%%%%%%%%%%%%%%
% \bibliographystyle{abbrv-no-doi}
\bibliographystyle{ieee}
\bibliography{ref}

%%%%%%%%%%%%%%%%%%%%%%%%%%%%%%%%% bio %%%%%%%%%%%%%%%%%%%%%%

\begin{IEEEbiography}[{\includegraphics[width=1in,height=1.25in,clip,keepaspectratio]{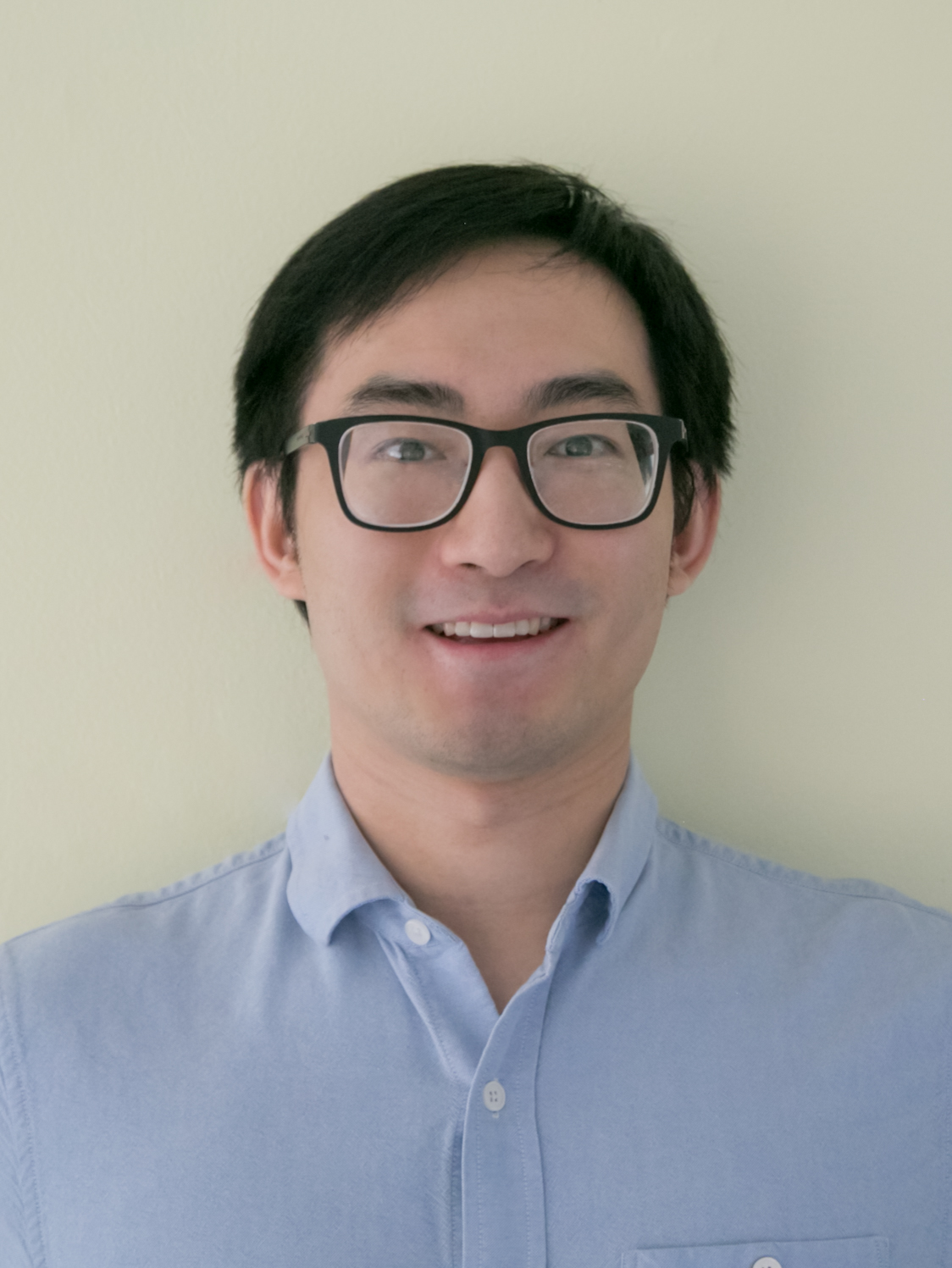}}]{Bingyao Huang}
  received the B.S. degree in electrical engineering from Tongji University, Shanghai, China, in 2013, and M.S. degree from Rowan University, Glassboro, NJ, USA, in 2015. He is currently pursuing a Ph.D. degree in Computer Science at Stony Brook University. His current research interests include computer vision, computational photography, augmented/virtual reality and deep learning. 
\end{IEEEbiography}
\vskip 0pt plus -1fil

\begin{IEEEbiography}[{\includegraphics[width=1in,height=1.25in,clip,keepaspectratio]{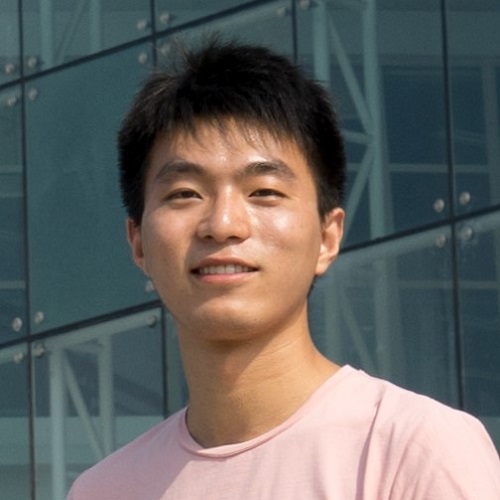}}]{Tao Sun}
  received the B.Eng. degree in automation from Huazhong University of Science and Technology, Wuhan, China, in 2015, and M.S. degree from Nanjing University, Nanjing, China, in 2018. He is currently pursuing a Ph.D. degree in Computer Science at Stony Brook University. His current research interests include computer vision, and machine learning. 
\end{IEEEbiography}
\vskip 0pt plus -1fil

\begin{IEEEbiography}[{\includegraphics[width=1in,height=1.25in,clip,keepaspectratio]{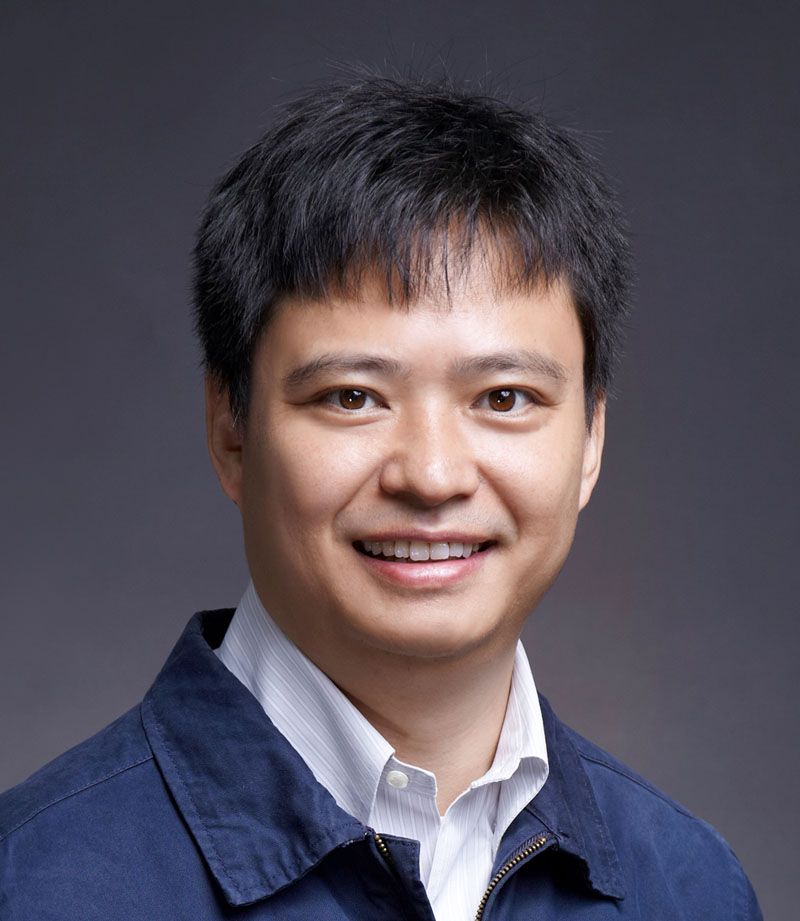}}]{Haibin Ling}
  received B.S. and M.S. from Peking University in 1997 and 2000, respectively, and Ph.D. from University of Maryland in 2006. From 2000 to 2001, he was an assistant researcher at Microsoft Research Asia; from 2006 to 2007, he worked as a postdoctoral scientist at UCLA; from 2007-2008, he worked for Siemens Corporate Research as a research scientist; and from 2008 to 2019, he was a faculty member of the Department of Computer Sciences for Temple University. In fall 2019, he joined the Department of Computer Science of Stony Brook University, where he is now a SUNY Empire Innovation Professor. His research interests include computer vision, augmented reality, medical image analysis, visual privacy protection, and human computer interaction. He received Best Student Paper Award of ACM UIST in 2003 and NSF CAREER Award in 2014. He serves as associate editors for IEEE Trans. on Pattern Analysis and Machine Intelligence (PAMI), Pattern Recognition (PR), and Computer Vision and Image Understanding (CVIU). He has served as Area Chairs various times for CVPR and ECCV.
\end{IEEEbiography}

\end{document}